%% file: main.tex
\documentclass[10pt,twocolumn,letterpaper]{article}
\usepackage{iccv}
\usepackage{times}
\usepackage{amsmath}
\usepackage{amssymb}
\usepackage{arydshln}
\usepackage{bm}
\usepackage{boldline}
\usepackage{booktabs}
\usepackage{color}
\usepackage{enumitem}
\usepackage{float}
\usepackage{graphicx}
\usepackage{multirow}
\usepackage{nicefrac}
\usepackage{subfig}
\usepackage[accsupp]{axessibility}
\usepackage[pagebackref=true,breaklinks=true,colorlinks,bookmarks=false]{hyperref}

\usepackage[capitalize]{cleveref}
\crefname{section}{Sec.}{Secs.}
\Crefname{section}{Section}{Sections}
\Crefname{table}{Table}{Tables}
\crefname{table}{Tab.}{Tabs.}

\makeatletter
\renewcommand{\paragraph}{%
  \@startsection{paragraph}{4}%
  {\z@}{0.25em}{-1em}%
  {\normalfont\normalsize\bfseries}%
}
\makeatother

\iccvfinalcopy
\def\iccvPaperID{1726}
\ificcvfinal\fi

\makeatletter
\def\input@path{{.}{..}}
\makeatother
\graphicspath{{.}{..}}

\def\mname{CVQ-VAE\xspace}

\title{Online Clustered Codebook}

\author{
Chuanxia Zheng \quad Andrea Vedaldi \\[0.3em]
Visual Geometry Group, University of Oxford \\
{\tt\small \{cxzheng, vedaldi\}@robots.ox.ac.uk}}

\begin{document}
\maketitle 
\ificcvfinal\thispagestyle{empty}\fi

\begin{abstract}
Vector Quantisation (VQ) is experiencing a comeback in machine learning, where it is increasingly used in representation learning.
However, optimizing the codevectors in existing VQ-VAE is not entirely trivial.
A problem is codebook collapse, where only a small subset of codevectors receive gradients useful for their optimisation, whereas a majority of them simply ``dies off'' and is never updated or used.
This limits the effectiveness of VQ for learning larger codebooks in complex computer vision tasks that require high-capacity representations.
In this paper, we present a simple alternative method for online codebook learning, Clustering VQ-VAE (\mname).
Our approach selects encoded features as anchors to update the ``dead'' codevectors, while optimising the codebooks which are alive via the original loss.
This strategy brings unused codevectors closer in distribution to the encoded features, increasing the likelihood of being chosen and optimized.
We extensively validate the generalization capability of our quantiser on various datasets, tasks (\eg reconstruction and generation), and architectures (\eg VQ-VAE, VQGAN, LDM).
\mname can be easily integrated into the existing models with just a few lines of code.
\end{abstract}

\begin{figure}[tb!]
    \centering
    \includegraphics[width=\linewidth]{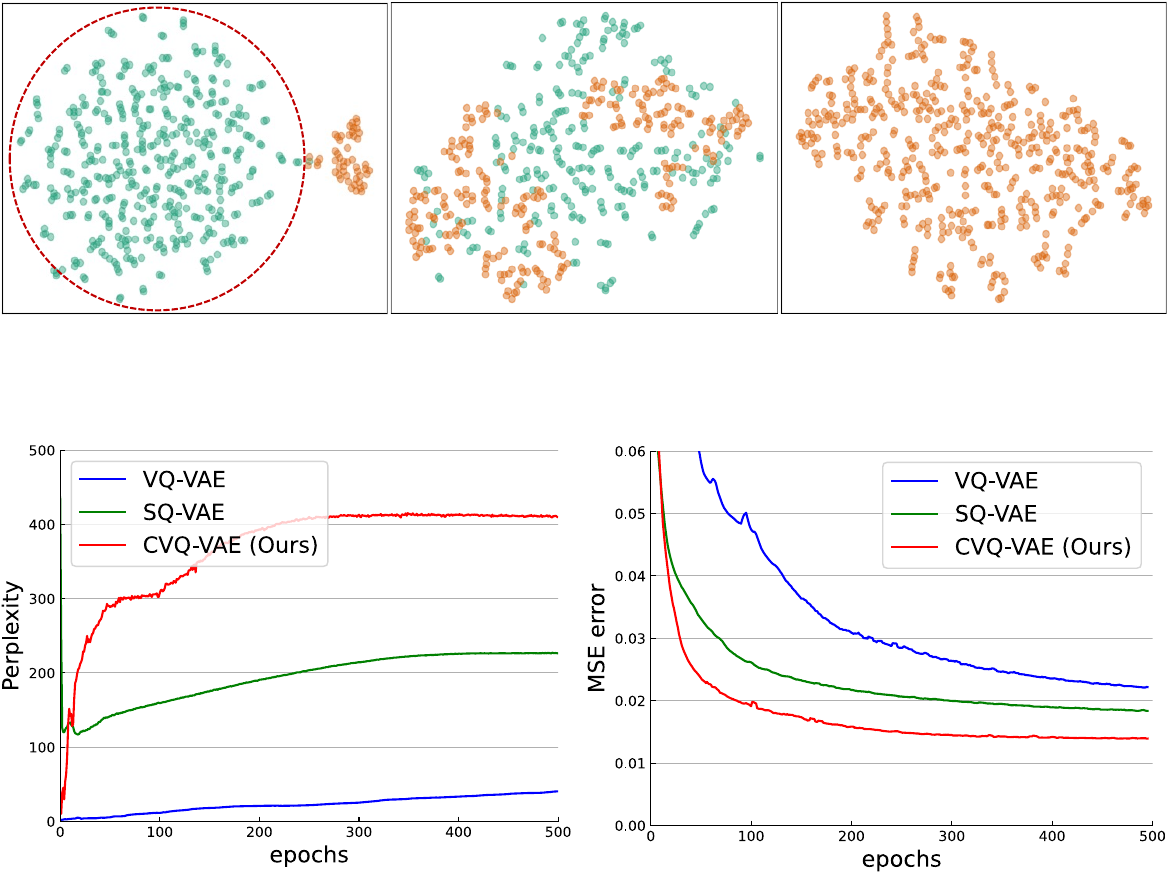}
    \begin{picture}(0,0)
    \put(-102,117){\footnotesize (a) VQ-VAE~\cite{van2017neural}}
    \put(-24,117){\footnotesize (b) SQ-VAE~\cite{takida2022sq}}
    \put(56,117){\footnotesize (c) \mname}
    \put(-97,106){\footnotesize Usage: $9.96\%$}
    \put(-24,106){\footnotesize Usage: $49.02\%$}
    \put(58,106){\footnotesize Usage: $100\%$}
    \put(-92,6){\footnotesize (d) Codebook Perplexity}
    \put(28,6){\footnotesize (e) Reconstruction error}
    \end{picture}
    \vspace{-0.3cm}
    \caption{\textbf{Codebook usage and reconstruction error.}
    The setting is the same as VQ-VAE~\cite{van2017neural}, except for the different quantisers.
    All models are trained and evaluated on the CIFAR10~\cite{krizhevsky2009learning} dataset.
    VQ-VAE has many ``dead'' vectors (\textcolor[RGB]{100,210,180}{green points}) which are \emph{not} used.
    \mname updates these unoptimized vectors by using online sampled feature anchors, leading to a 100\% usage of the codebook.
    \mname achieves substantially higher codebook perplexity and better reconstruction results than with the fixed initialization.}%
    \label{fig:intro_mov}
\end{figure}
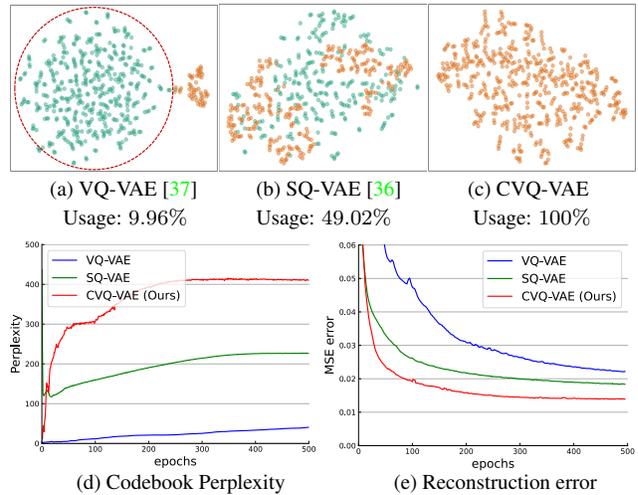

\section{Introduction}

Vector Quantisation (VQ)~\cite{1162229} is a basic building block of many machine learning techniques.
It is often used to help learning unsupervised representations for vision and language tasks, including data compression~\cite{agustsson2017soft,williams2020hierarchical,takida2022sq}, recognition~\cite{maodiscrete,baobeit,yu2022vectorquantized,liu2022cross,li2022unimo}, and generation~\cite{van2017neural,razavi2019generating,esser2021taming,rombach2022high,zhengmovq,sargent2023vq3d,sanghi2023sketch}.
VQ quantises continuous feature vectors into a discrete space by embedding them to the closest vectors in a codebook of representatives or codevectors.
Quantisation has been shown to simplify optimization problems by reducing a continuous search space to a discrete one.

Despite its success, VQ has some drawbacks when applied to deep networks~\cite{van2017neural}.
One of them is that quantisation stops gradients from back-propagating to the codevectors.
This has been linked to \emph{codebook collapse}~\cite{takida2022sq}, which means that only a small subset of active codevectors are optimized alongside the learnable features, while the majority of them \emph{are not used at all} (see the \textcolor[RGB]{100,210,180}{green} ``dead'' points in \cref{fig:intro_mov}(a)).
As a result, many recent methods~\cite{esser2021taming,esser2021imagebart,yu2022vectorquantized,rombach2022high,chang2022maskgit,zhengmovq} fail to utilise the full expressive power of a codebook due to the low codevector utilisation, especially when the codebook size is large.
This significantly limits VQ's effectiveness.

\begin{figure*}[tb!]
    \centering
    \includegraphics[width=\linewidth]{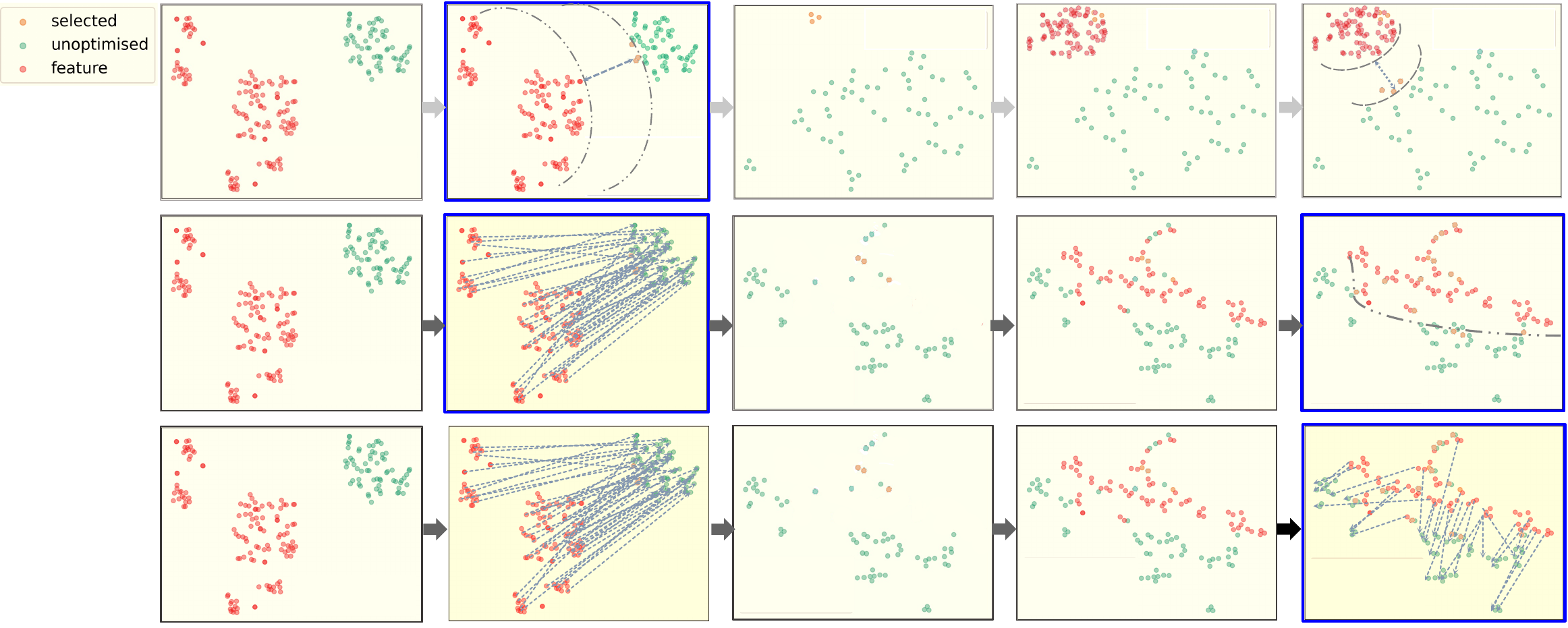}
    \begin{picture}(0,0)
    \put(-243,175){\footnotesize VQ-VAE~\cite{van2017neural}}
    \put(-240,110){\footnotesize \mname }
    \put(-235,100){\footnotesize (offline)}
    \put(-240,44){\footnotesize \mname}
    \put(-235,34){\footnotesize (online)}
    \put(-180,5){\footnotesize (a1) Distribution}
    \put(-102,5){\footnotesize (b1) Code vector update}
    \put(-11,5){\footnotesize (c1) Updated codebook}
    \put(91,5){\footnotesize (a2) Distribution}
    \put(170,5){\footnotesize (b2) Code vector update}
    \end{picture}
    \vspace{-0.3cm}
    \caption{\textbf{Codebook optimization}.
    The \textcolor[RGB]{230,128,128}{Red points} indicate the encoded features, while the \textcolor[RGB]{100,210,180}{Green} and \textcolor[RGB]{240,190,140}{Peach points} denote the unoptimized and active vectors in the codebook, respectively.
    1) In VQ-VAE~\cite{van2017neural} (row 1), only the active ``lucky'' seeds (in \textcolor[RGB]{240,190,140}{Peach}) are optimized alongside the encoded features (in \textcolor[RGB]{230,128,128}{Red}) during training. The other ``dead'' vectors (in \textcolor[RGB]{100,210,180}{Green}) are \emph{not} given attention and remain fixed.
    2) In our \mname(offline) (row 2), we reinitialize the codevectors based on the anchors sampled from the encoded features (in \textcolor[RGB]{230,128,128}{Red}), encouraging the ``dead'' ones to be closer to the features in distribution.
    3) To address the difficulty of covering all samples by single sampling in mini-batch learning, we further propose an online learning variant (row 3), where the anchor is obtained by calculating the moving average of the encoded features in different batches.
    We highlighted the difference between various methods in the blue thickened border.}
    \label{fig:method}
\end{figure*}
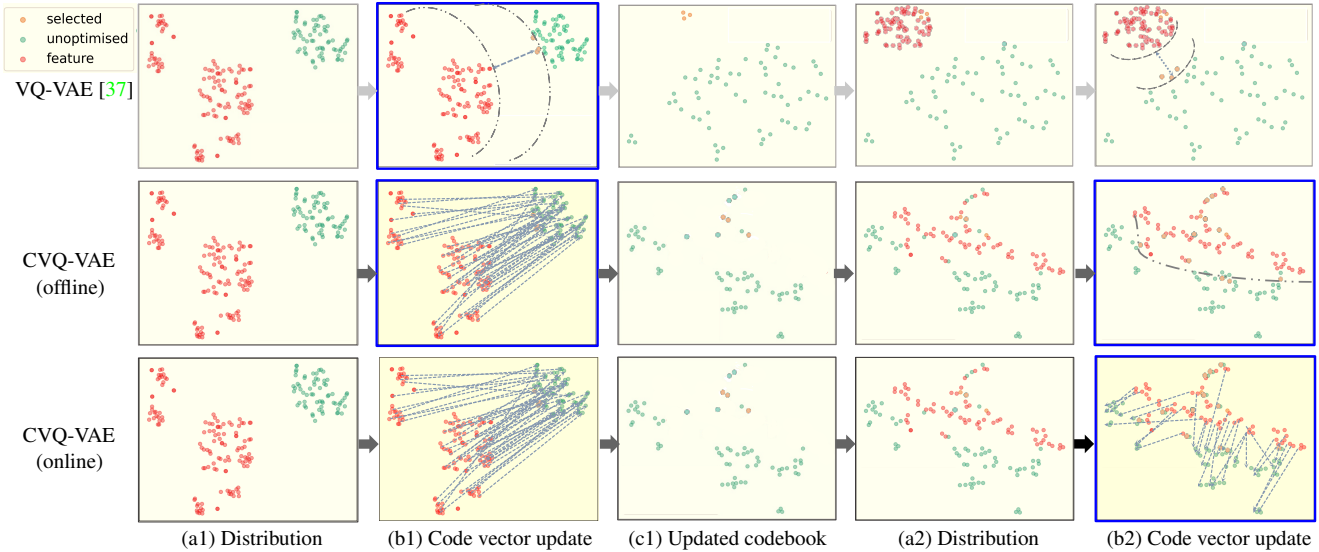

To tackle this issue, we propose a new alternative quantiser called \emph{Clustering VQ-VAE} (\mname).
We observe that classical clustering algorithms, such as refined initialization $k$-means~\cite{bradley1998refining} and $k$-means++~\cite{arthur2007k}, use a dynamic cluster initialization approach.
For example, $k$-means++ randomly selects a data point as the first cluster centre, and then chooses the next new centre based on a weighted probability calculated from the distance to the previous centres.
Analogously, \mname \emph{dynamically} initializes unoptimized codebooks by resampling them from the learned features (\cref{fig:method}).
This simple approach can avoid codebook collapse and significantly enhance the usage of larger codebooks by enabling optimization of all codevectors (achieving $100\%$ codebook utilisation in \cref{fig:intro_mov}{(c)}).

While \mname is inspired by previous dynamic cluster initialization techniques~\cite{bradley1998refining,arthur2007k}, its implementation in deep networks requires careful consideration.
Unlike traditional clustering algorithms~\cite{lloyd1982least,bradley1998refining,hamerly2002alternatives,arthur2007k} where source data points are fixed, in deep networks features and their corresponding codevectors are mutually and incrementally optimized.
Thus, simply sampling codevectors from a single snapshot of features would not work well because any mini-batch used for learning \emph{cannot} capture the true data distribution, as demonstrated in our offline version in \cref{tab:rule}.
To fix this issue, we propose to \emph{compute running averages} of the encoded features across different training mini-batches and use these to improve the dynamic reinitialization of the collapsed codevectors.
This operation is similar to an online feature clustering method that calculates average features across different training iterations (\cref{fig:method}).
While this may seem a minor change, it leads to a very significant improvement in terms of performance (\cref{fig:intro_mov}{(e)}).

As a result of these changes, \mname significantly outperforms the previous models VQ-VAE~\cite{van2017neural} and SQ-VAE~\cite{takida2022sq} on various datasets under the same setting, and with no other changes except for swapping in the new quantiser.
Moreover, we conduct thorough ablation experiments on variants of the method to demonstrate the effectiveness of our design and analyse the importance of various design factors.
Finally, we incorporate \mname into large models (\eg VQ-GAN~\cite{esser2021taming} and LDM~\cite{rombach2022high}) to further demonstrate its generality and potential in various applications.

\section{Related Works}

VQ-VAE~\cite{van2017neural} learns to quantise the continuous features into a discrete space using a restricted number of codebook vectors.
By clustering features in the latent space, VQ-VAE can automatically learn a crucially compact representation and store the domain information in the decoder that does not require supervision.
This discrete representation has been applied to various downstream tasks, including image generation~\cite{razavi2019generating,yu2022vectorquantized,chang2022maskgit,lee2022autoregressive,hu2022global}, image-to-image translation~\cite{esser2021taming,ramesh2021zero,esser2021imagebart,rombach2022high}, text-to-image synthesis~\cite{ramesh2021zero,ding2021cogview,ramesh2022hierarchical,hu2022unified}, conditional video generation~\cite{rakhimov2021latent,wu2022nuwa,yan2021videogpt}, image completion~\cite{esser2021taming,esser2021imagebart,zheng2022high}, recognition~\cite{maodiscrete,baobeit,yu2022vectorquantized,liu2022cross,li2022unimo} and 3D reconstruction~\cite{mittal2022autosdf,sargent2023vq3d,sanghi2023sketch}.

Among them, VQ-GAN~\cite{esser2021taming}, ViT-VQGAN~\cite{yu2022vectorquantized}, RQ-VAE~\cite{lee2022autoregressive}, and MoVQ~\cite{zheng2022high} aim to train a better discrete representation through deeper network architectures, additional loss functions, multichannel or higher resolution representations. However, none of them tackle the \emph{codebook collapse} issue for the unoptimized ``dead'' point.

To address this issue, additional training heuristics are proposed in recent works. 
SQ-VAE~\cite{takida2022sq} improves VQ-VAE with stochastic quantisation and a trainable posterior categorical distribution. 
VQ-WAE~\cite{vuong2023vector} builds upon SQ-VAE by directly encouraging the discrete representation to be a uniform distribution via a \emph{Wasserstein} distance. 
The most related works are HVQ-VAE~\cite{williams2020hierarchical} and Jukebox~\cite{dhariwal2020jukebox} that use \emph{codebook reset} to randomly reinitialize unused or low-used codebook entries. 
However, they only assign a single sampled anchor to each unoptimized codevector. 
In contrast, our \mname considers the changing of features in deep networks and designs an online clustering algorithm by running average updates across the training mini-batch. 
Additionally, our work bridges codebook reset in Jukebox for music generation to the more general class of running average updates that are applicable to image compression and generation problems in computer vision.

\section{Method}%
\label{s:method}

VQ is in the context of unsupervised representation learning.
Our main goal is to learn a discrete codebook that efficiently utilizes \emph{all codebook entries within it}.
To achieve this, our quantisation method, as illustrated in \cref{fig:method}, is conceptually similar to VQ-VAE~\cite{van2017neural}, except that our codevectors are \emph{dynamically initialized} rather than being sampled from a \emph{fixed} uniform or Gaussian distribution.
In the following sections, we provide a general overview of VQ (\cref{sec:method_back}), followed by 
our proposed \mname (\cref{sec:mth_online}).

\subsection{Background: VQ-VAE}%
\label{sec:method_back}

Given a high dimensional image $x\in\mathbb{R}^{H\times W\times c}$, VQ-VAE~\cite{van2017neural} learns to embed it with low dimensional codevectors $z_q\in\mathbb{R}^{h \times w\times n_q}$, where $n_q$ is the dimensionality of the vectors in the codebook.
Then, the feature tensor can be equivalently described as a compact representation with $h\times w$ indices corresponding to the codebook entries $z_q$.
This is done via an autoencoder
\begin{equation}\label{eq:vq-vae}
    \hat{x}
    = \mathcal{G}_\theta(z_q)
    = \mathcal{G}_\theta(\mathbf{q}(\hat{z}))
    = \mathcal{G}_\theta(\mathbf{q}(\mathcal{E}_\phi(x))).
\end{equation}
Here $\mathcal{E}_\phi$ and $\mathcal{G}_\theta$ refer to the encoder and decoder, respectively.
The encoder embeds images into the continuous latent space, while the decoder inversely maps the latent vectors back to the original image.
$\mathbf{q}(\cdot)$ is a quantisation operation that maps the continuous encoded observations $\hat{z}$ into the discrete space by looking up the closest codebook entry $e_k$ for each grid feature $\hat{z}_i$ using the following equation:
\begin{equation}\label{eq:quant}
    z_{q_{i}}
    = \mathbf{q}(\hat{z}_{i})
    = e_k,\quad\text{where}\quad
    k
    = \underset{e_k\in\mathcal{Z}}{\operatorname{argmin}}\lVert\hat{z}_{i}-e_k\rVert,
\end{equation}
where $\mathcal{Z}=\{e_k\}_{k=1}^K$ is the codebook that consists of $K$ entries $e_k\in\mathbb{R}^{n_q}$ with dimensionality $n_q$.
During training, the encoder $\mathcal{E}_\phi$, decoder $\mathcal{G}_\theta$ and codebook $\mathcal{Z}$ are jointly optimized by minimizing the following objective:
\begin{equation}\label{eq:vae_loss}
    \mathcal{L}
    = \lVert x - \hat{x} \rVert_2^2 + \lVert\mathrm{sg}[\mathcal{E}_\psi(x)] - z_q\rVert_2^2 + \beta\lVert\mathcal{E}_\psi(x) - \mathrm{sg}[z_q]\rVert_2^2,
\end{equation}
where $\mathrm{sg}$ denotes a stop-gradient operator, and $\beta$ is the hyperparameter for the last term \emph{commitment loss}.
The first term is known as \emph{reconstruction loss}, which measures the difference between the observed $x$ and the reconstructed $\hat{x}$.
The second term is the \emph{codebook loss}, which encourages the codevectors to be close to the encoded features.
In practice, the codebook $\mathcal{Z}$ is optimized using either the \emph{codebook loss}~\cite{van2017neural} or using an exponential moving average (EMA)~\cite{razavi2019generating}.
However, these methods work only for the active codevectors, \emph{leaving the ``dead'' ones unoptimized}.

\subsection{Clustering VQ-VAE (CVQ-VAE)}%
\label{sec:mth_online}

The choice of initial points is a crucial aspect of unsupervised codebook learning.
Classical clustering methods like refined $k$-means~\cite{bradley1998refining} and $k$-means++~\cite{arthur2007k} are \emph{dynamically-initialized}, which means that each new clustering centre is initialized based on previously calculated distance or points.
This leads to a more robust and effective clustering result, as reported in comparative studies~\cite{celebi2013comparative}.

Analogously, we build a \emph{dynamically-initialized} vector quantized codebook in deep networks. 
However, unlike traditional clustering settings, the data points, \ie the encoded features $\hat{z}$ in the deep network, are also updated during training instead of being fixed.
Therefore, a dynamical initialization strategy should take into account the changing feature representations during training.

\paragraph{Running average updates.}

To build the online initialization for a codebook, we start by accumulatively counting the average usage of codevectors in each training mini-batch:
\begin{equation}\label{eq:count}
    N_k^{(t)} 
    = N_k^{(t-1)} \cdot \gamma + \frac{n_k^{(t)}}{Bhw} \cdot (1-\gamma),
\end{equation}
where $n_k^{(t)}$ is the number of encoded features in a training mini-batch that will be quantised to the closest codebook entry $e_k$, and $Bhw$ denotes the number of features on Batch, height, and width.
$\gamma$ is a decay hyperparameter with a value in $(0,1)$ (default $\gamma=0.99$).
$N_k^{(0)}$ is initially set as zero.

We then select a subset $\bar{\mathcal{Z}}$ with $K$ vectors from the encoded features $\hat{z}$, which we denote as \textbf{anchors}.
Instead of directly using the anchors to reinitialize the unoptimized codevectors, we expect that \emph{codevectors that are less-used or unused should be modified more than frequently used ones}.
To achieve this goal, we compute a decay value $a_k^{(t)}$ for each entry $e_k$ using the accumulative average usage $N_k^{(t)}$ and reinitialize the features as follows:
\begin{gather}\label{eq:update}
    \alpha_k^{(t)} 
    = \exp^{-N_k^{(t)} K\,\frac{10}{1-\gamma}-\epsilon}, \\
    e_k^{(t)} 
    = e_k^{(t-1)} \cdot (1-\alpha_k^{(t)}) + \hat{z}_k^{(t)} \cdot \alpha_k^{(t)},
\end{gather}
where $\epsilon$ is a small constant to ensure the entries are assigned with the average values of features along different mini-batches, and $\hat{z}_k^{(t)}\in{\bar{\mathcal{Z}}}^{K\times z_q}$ is the sampled anchor.

This running average operation differs from the exponential moving average (EMA) used in VQ-VAE~\cite{razavi2019generating}.
Our equation is applied to reinitialize unused or low-used codevectors, instead of updating the active ones.
Furthermore, our decay parameter in \cref{eq:update} is computed based on the average usage, which is \emph{not a pre-defined hyperparameter}.

\paragraph{Choice of the anchors.} Next, we describe several versions of the anchor sampling methods.
Interestingly, experimental results (\cref{tab:rec_ab_featm}) show that our online version is \emph{not} sensitive to these choices.
However, the different anchor sampling methods have a direct impact on the \emph{offline} version, suggesting that our running average updates behaviour is the primary reason for the observed improvements.
\begin{itemize}[itemsep=0pt]
    \vspace{-0.1cm}
    \item \textbf{Random.} Following the codebook reset~\cite{dhariwal2020jukebox,williams2020hierarchical}, a natural choice of anchors is randomly sampled from the encoded features.
    \item \textbf{Unique.} To avoid repeated anchors, a random permutation of integers within the number of features ($Bhw$) is performed. Then, we select the first $K$ features.
    \item \textbf{Closest.} A simple way is inversely looking up the closest features of each entry, \ie $i = \underset{\hat{z}_i\in\mathcal{E}_\phi(x)}{\operatorname{argmin}}\lVert\hat{z}_{i}-e_k\rVert$.
    \item \textbf{Probabilistic random.} We can also sample anchors based on the distance $D_{i,k}$ between the codevectors and the encoded features.
    In this paper, we consider the probability $p=\frac{\exp{(-D_{i,k})}}{\sum_{i=1}^{Bhw}\exp{(-D_{i,k})}}$.
\end{itemize}


\paragraph{Contrastive loss.}

We further introduce a contrastive loss
$
-\log
\frac
{e^{sim(e_k,\hat{z}_i^+)/\tau}}
{\sum_{i=1}^N e^{sim(e_k,\hat{z}_i^-)/\tau}}
$
to encourage sparsity in the codebook.
In particular, for each codevector $e_k$, we directly select the closest feature $\hat{z}_i^+$ as the positive pair and sample other farther features $\hat{z}_i^-$ as negative pairs using the $D_{i,k}$.

\paragraph{Relation to prior work.}

To mitigate the codebook collapse issue, several methods have been proposed, like 
normalized codevectors in ViT-VQGAN~\cite{yu2022vectorquantized}.
However, these methods only optimize the \emph{active} entries, rather than the entire codebook.
Recently, SQ-VAE~\cite{takida2022sq}, SeQ-GAN~\cite{gu2022rethinking}, and VQ-WAE~\cite{vuong2023vector} assume that the codebook follows a fixed distribution.
Although these methods achieve high perplexity, the reconstruction quality is \emph{not} always improved (\cref{tab:rec_ab}). 
The most relevant work to ours is codebook reset, which randomly reinitializes the unused or low-used codevectors to high-usage ones~\cite{williams2020hierarchical} or encoder outputs~\cite{dhariwal2020jukebox}.
However, these methods rely only on a temporary single value for initialization and miss the opportunity of exploiting online clustering across different training steps.

\section{Experiments: Image Quantisation}%
\label{sec:exp}

\subsection{Experimental Details}

\paragraph{Implementation.} \mname can be easily implemented by a few lines of code in Pytorch, where the gradient for the selected codevectors is preserved. The code is available at \href{https://github.com/lyndonzheng/CVQ-VAE}{https://github.com/lyndonzheng/CVQ-VAE}.

Our implementation is built upon existing network architectures.
We set all hype-parameters following the original code, except that we replace these quantisers with our online clustering codebook.
In particular, we first demonstrate our assumption on small datasets with the officially released VQ-VAE~\cite{van2017neural} implementation~\footnote{\href{https://github.com/deepmind/sonnet/blob/v2/sonnet/src/nets/vqvae.py}{https://github.com/deepmind/sonnet/blob/v2/sonnet/src/nets/vqvae.py}}$^,$\footnote{\href{https://github.com/deepmind/sonnet/blob/v1/sonnet/examples/vqvae_example.ipynb}{https://github.com/deepmind/sonnet/blob/v1/sonnet/examples/vqvae\_
example.ipynb}}. 
Then, we verify the generality of our quantiser on large datasets using the officially released VQ-GAN~\cite{esser2021taming} architecture~\footnote{\href{https://github.com/CompVis/taming-transformers}{https://github.com/CompVis/taming-transformers}}.

\begin{figure*}[tb!]
    \centering
    \includegraphics[width=\linewidth]{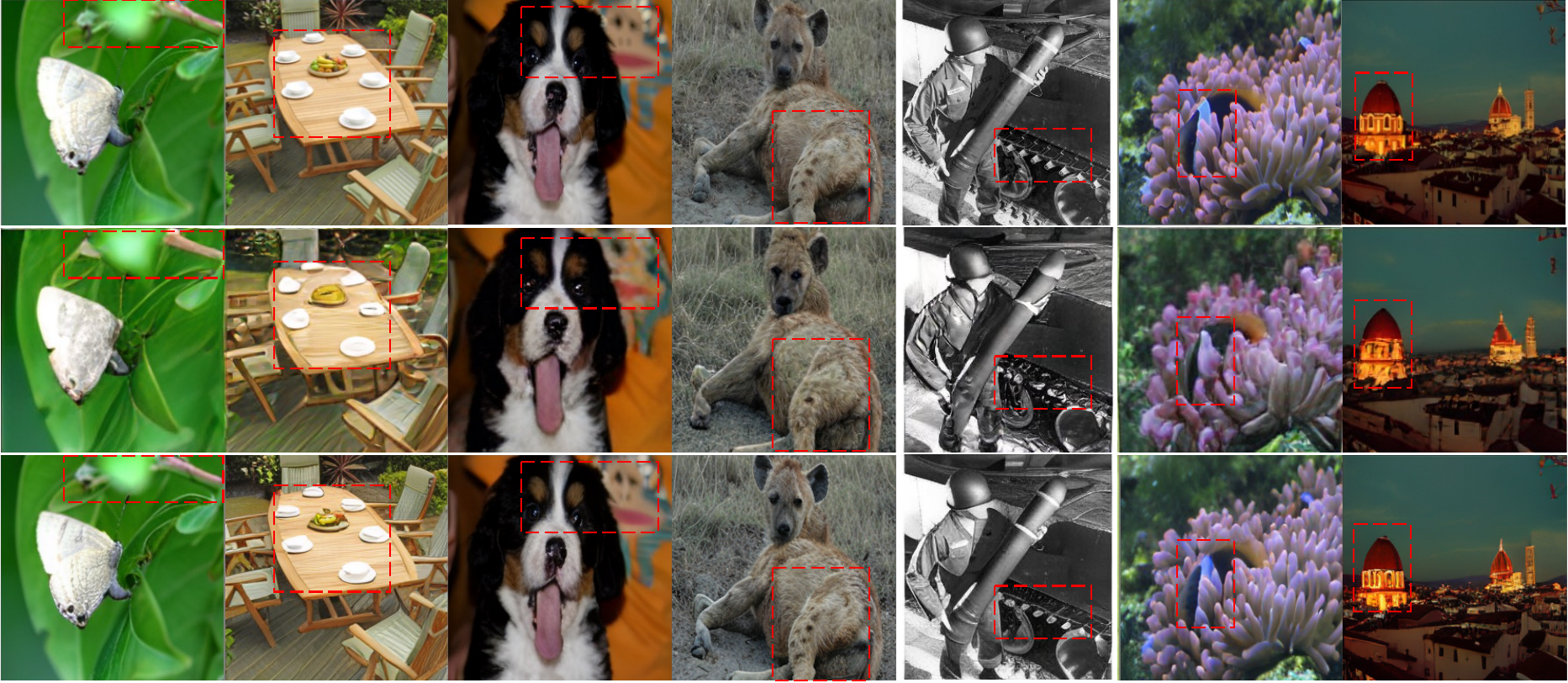}
    \begin{picture}(0,0)
    \put(-256,190){\rotatebox{90}{\footnotesize Input}}
    \put(-256,95){\rotatebox{90}{\footnotesize VQGAN~\cite{esser2021taming}}}
    \put(-256,26){\rotatebox{90}{\footnotesize \mname}}
    \end{picture}
    \vspace{-0.6cm}
    \caption{\textbf{Reconstructions from different models.} The two models are trained under the same settings, except for the different quantisers. Compared with the state-of-the-art baseline VQGAN~\cite{esser2021taming}, the proposed model significantly improves the reconstruction quality (highlight in \textcolor[RGB]{210,0,0}{red box}) under the same compression ratio (768$\times$, with 16$\times$ downsampling).}%
    \label{fig:rec}
    \vspace{-0.3cm}
\end{figure*}
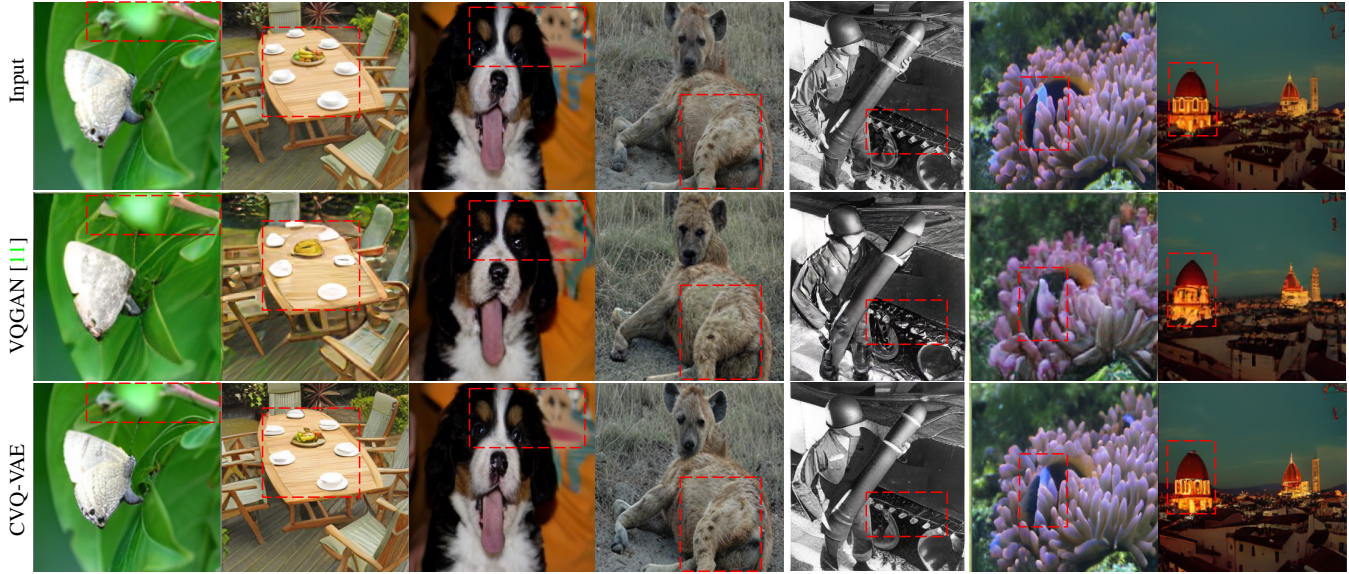

\paragraph{Datasets.}

We evaluated the proposed quantiser on various datasets, including MNIST~\cite{lecun1998gradient}, CIFAR10~\cite{krizhevsky2009learning}, Fashion MNIST~\cite{xiao2017online}, and the higher-resolution FFHQ~\cite{karras2019style} and the large ImageNet~\cite{deng2009imagenet}.

\paragraph{Metrics.}

Following existing works~\cite{esser2021taming,zhengmovq,gu2022rethinking}, we evaluated the image quality between reconstructed and original images on different scales, including patch-level structure similarity index (SSIM), feature-level Learned Perceptual Image Patch Similarity (LPIPS)~\cite{zhang2018unreasonable}, and dataset-level Fr\'{e}chet Inception Distance (FID)~\cite{heusel2017gans}.
We also report the perplexity score for the codebook ablation study as in SQ-VAE~\cite{takida2022sq} and VQ-WAE~\cite{vuong2023vector}.
It is defined as
$
e^{-\sum_{k=1}^K p_{e_k}\log p_{e_k}}
$,
where $p_{e_k}=\frac{n_k}{\sum_{i=i}^K n_k}$, and $n_k$ is the number of encoded features associated with codevector $e_k$.

\subsection{Main Results}

\begin{table}[tb!]
    \centering
    \renewcommand{\arraystretch}{1.0}
    \setlength\tabcolsep{5pt}
    \begin{tabular}{@{}l c ccc @{}}
    \toprule
    \textbf{Method} & \textbf{Dataset} & SSIM $\uparrow$ & LPIPS $\downarrow$ & rFID $\downarrow$ \\
    \midrule
    VQ-VAE~\cite{van2017neural} & \multirow{4}{*}{MNIST}& 0.9777 & 0.0282 & 3.43 \\
    HVQ-VAE~\cite{williams2020hierarchical} & & 0.9790 & 0.0270 & 3.17 \\
    SQ-VAE~\cite{takida2022sq} & & 0.9819 & 0.0256 & 3.05 \\
    \cdashline{1-5}
    \textbf{\mname} & & \textbf{0.9833} & \textbf{0.0222} & \textbf{1.80} \\
    \midrule
    VQ-VAE~\cite{van2017neural} & \multirow{4}{*}{CIFAR10}& 0.8595 & 0.2504 & 39.67 \\
    HVQ-VAE~\cite{williams2020hierarchical} & & 0.8553 & 0.2553 & 41.08 \\
    SQ-VAE~\cite{takida2022sq} & & 0.8779 & 0.2333 & 37.92 \\
    \cdashline{1-5}
    \textbf{\mname} & & \textbf{0.8978} & \textbf{0.1883} & \textbf{24.73} \\
    \bottomrule
    \end{tabular}
    \caption{\textbf{Reconstruction results} on the validation sets of MNIST (10,000 images) and CIFAR10 (10,000 images). All models are trained with the same experimental settings, except for the different quantisers.}%
    \label{tab:rec_quant}
\end{table}

\begin{table}[tb!]
    \centering
    \renewcommand{\arraystretch}{1.0}
    \setlength\tabcolsep{2pt}
    \begin{tabular}{@{}l c cccc @{}}
    \toprule
         \textbf{Method} &  \textbf{Dataset} & $\mathcal{S}$ $\downarrow$ & $\mathcal{K}$ $\downarrow$ & Usage $\uparrow$ & rFID $\downarrow$ \\
    \midrule
    VQGAN~\cite{esser2021taming} & \multirow{6}{*}{\rotatebox{90}{FFHQ}}& 16$^2$& 1024& $42\%$& 4.42\\
    ViT-VQGAN~\cite{yu2022vectorquantized} & & 32$^2$& 8192& --- & 3.13\\
    RQ-VAE~\cite{lee2022autoregressive} & & 16$^2\times$4 & 2048 & --- & 3.88 \\
    MoVQ~\cite{zhengmovq} & & 16$^2\times$4 & 1024 & $56\%$& 2.26$^*$ \\
    SeQ-GAN~\cite{gu2022rethinking} & & 16$^2$& 1024 & $100\%$& 3.12 \\
    \cdashline{1-6}
    \textbf{\mname} (ours) & & 16$^2$& 1024 & $100\%$ & \underline{2.80}\\
    \textbf{\mname} (ours) & & 16$^2\times$4 & 1024 & $100\%$ & \textbf{2.03}\\
    \midrule
    VQGAN~\cite{esser2021taming} & \multirow{6}{*}{\rotatebox{90}{ImageNet}}& 16$^2$& 1024 & $44\%$ & 7.94 \\
    ViT-VQGAN~\cite{yu2022vectorquantized} & & 32$^2$& 8192 & $96\%$& \underline{1.28} \\
    RQ-VAE~\cite{lee2022autoregressive} & & 8$^2\times$16& 16384 & --- & 1.83\\
    MoVQ~\cite{zhengmovq} & & 16$^2\times$4& 1024 & $63\%$& \textbf{1.12} \\
    SeQ-GAN~\cite{gu2022rethinking} & & 16$^2$ & 1024 & $100\%$ & 1.99\\
    \cdashline{1-6}
    \textbf{\mname} (ours) & & 16$^2$& 1024 & $100\%$ & 1.57 \\
    \textbf{\mname} (ours) & & 16$^2\times$4 & 1024 & $100\%$ & 1.20$^*$ \\
    \bottomrule
    \end{tabular}
    \caption{\textbf{Reconstruction results} on validation sets of ImageNet (50,000 images) and FFHQ (10,000 images). $\mathcal{S}$ denotes the latent size of encoded features, and $\mathcal{K}$ is the number of codevectors in the codebook. Usage indicates how many entries in a codebook are used during the quantisation on the validation set. More evaluation metrics are reported in Appendix Table \ref{tab:app_rec_sota}.}%
    \label{tab:rec_sota}
\end{table}

\begin{table*}[tb!]
    \centering
    \renewcommand{\arraystretch}{1.0}
    \setlength\tabcolsep{3pt}
    \begin{tabular}{@{}ll cccc cccc cccc@{}}
    \toprule
         &  \multirow{2}{*}{\textbf{Method}} & \multicolumn{3}{c}{\textbf{MNIST} (28$\times$28)} && \multicolumn{3}{c}{\textbf{CFAIR10} (32$\times$32)} && \multicolumn{3}{c}{\textbf{Fashion MNIST} (28$\times$28)}\\
         \cline{3-5}\cline{7-9}\cline{11-13}
         &  & SSIM $\uparrow$ & LPIPS $\downarrow$ & rFID $\downarrow$ && SSIM $\uparrow$ & LPIPS $\downarrow$ & rFID $\downarrow$ && SSIM $\uparrow$ & LPIPS $\downarrow$ & rFID $\downarrow$\\
    \midrule
    $\mathbb{A}$ & Baseline VQ-VAE~\cite{van2017neural}$_{\text{\scriptsize{NeurIPS'2017}}}$ & 0.9777 & 0.0282 & 3.43 && 0.8595 & 0.2504 & 39.67 && 0.9140 & 0.0801 & 12.73 \\
    \cdashline{1-13}
    $\mathbb{B}$ & + Cosine distance & 0.9791 & 0.0266 & 3.06 && 0.8709 & 0.2303 & 35.14 && 0.9160 & 0.0764 & 11.40 \\
    $\mathbb{C}$ & + Anchor initialization (offline) &  0.9810 & 0.0253 & 2.78 && 0.8829 & 0.2132 & 31.10 && 0.9145 & 0.0773 & 11.92 \\
    $\mathbb{D}$ & + Anchor initialization (online) & 0.9823 & 0.0236 & 2.23 && \textbf{0.8991} & 0.1897 & 26.62  && \textbf{0.9254} & \textbf{0.0683} & 9.27 \\
    $\mathbb{E}$ & + Contrastive loss &  \textbf{0.9833} & \textbf{0.0222} & \textbf{1.80} && 0.8978 & \textbf{0.1883} & \textbf{24.73} && 0.9233 & 0.0693 & \textbf{8.85}\\
    \bottomrule
    \end{tabular}
    \vspace{-0.1cm}
    \caption{\textbf{Results on various settings.} We report patch-level SSIM, feature-level LPIPS, and dataset-level FID\@. All evaluation metrics are reported in Appendix Table \ref{tab:app_rule}.}%
    \vspace{-0.3cm}
    \label{tab:rule}
\end{table*}

\paragraph{Quantitative Results:} We first evaluated our \mname and various quantisers, including VQ-VAE~\cite{van2017neural}$_{\text{\scriptsize{NeurIPS'2017}}}$, HVQ-VAE~\cite{williams2020hierarchical}$_{\text{\scriptsize{NeurIPS'2020}}}$, and SQ-VAE~\cite{takida2022sq}$_{\text{\scriptsize{ICML'2022}}}$, under the identical experimental settings in \cref{tab:rec_quant}. 
All instantiations of our model outperform the baseline variants of previous state-of-the-art models. 
Although the latest SQ-VAE~\cite{takida2022sq} optimizes all code entries by explicitly enforcing the codebook to be a defined distribution, this assumption may not hold for all datasets.
For instance, code entries that respond to the background elements like sky and ground should take more count than code entries that represent specific objects, such as vehicle wheels.
In contrast, our quantiser only encourages all code entries to be optimized, leaving the association to be automatically learned.

Then, we compared our \mname with the state-of-the-art methods, including VQGAN \cite{esser2021taming}$_{\text{\scriptsize{CVPR'2021}}}$, ViT-VQGAN \cite{yu2022vectorquantized}$_{\text{\scriptsize{ICLR'2022}}}$, RQ-VAE \cite{lee2022autoregressive}$_{\text{\scriptsize{CVPR'2022}}}$, and MoVQ \cite{zhang2018unreasonable}$_{\text{\scriptsize{NeurIPS'2022}}}$ for the task of reconstruction.
\Cref{tab:rec_sota} shows quantitative results on two large datasets.
Under the same compression ratio (768$\times$, \ie 256$\times$256$\times$3$\to$16$\times$16), our model significantly outperforms the state-of-the-art models, including the baseline VQGAN~\cite{esser2021taming} and the concurrent SeQ-GAN~\cite{gu2022rethinking}.
Interestingly, on the FFHQ dataset, our model even outperforms ViT-VQGAN~\cite{yu2022vectorquantized} and RQ-VAE~\cite{lee2022autoregressive}, which utilize 4$\times$ tokens for the representation.
This suggests that the high usage of codevectors is significant for maintaining information during data compression.
Additionally, we also run $4\times$ tokens experiments, as in MoVQ~\cite{zhengmovq}. The \mname further achieves a relative 10.1\% improvement.
Although our 4$\times$ version shows a slightly lower rFID score than the MoVQ~\cite{zhengmovq} on ImageNet dataset (1.12 \emph{vs.} 1.20), we achieve better performance on other metrics (as shown in Appendix \cref{tab:app_rec_sota}).

\begin{table*}[tb!]
    \centering
    \subfloat[
    \textbf{Codebook size}. The blobs' size is proportional to the number of codebook vectors \{32, 64, 128, 256, 512,  1024\}. The larger size naturally leads to better results in our \mname.
    \label{tab:rec_ab_codesize}
    ]{
    \centering
    \begin{minipage}{0.30\linewidth}
    {
    \begin{center}
        \includegraphics[width=\linewidth]{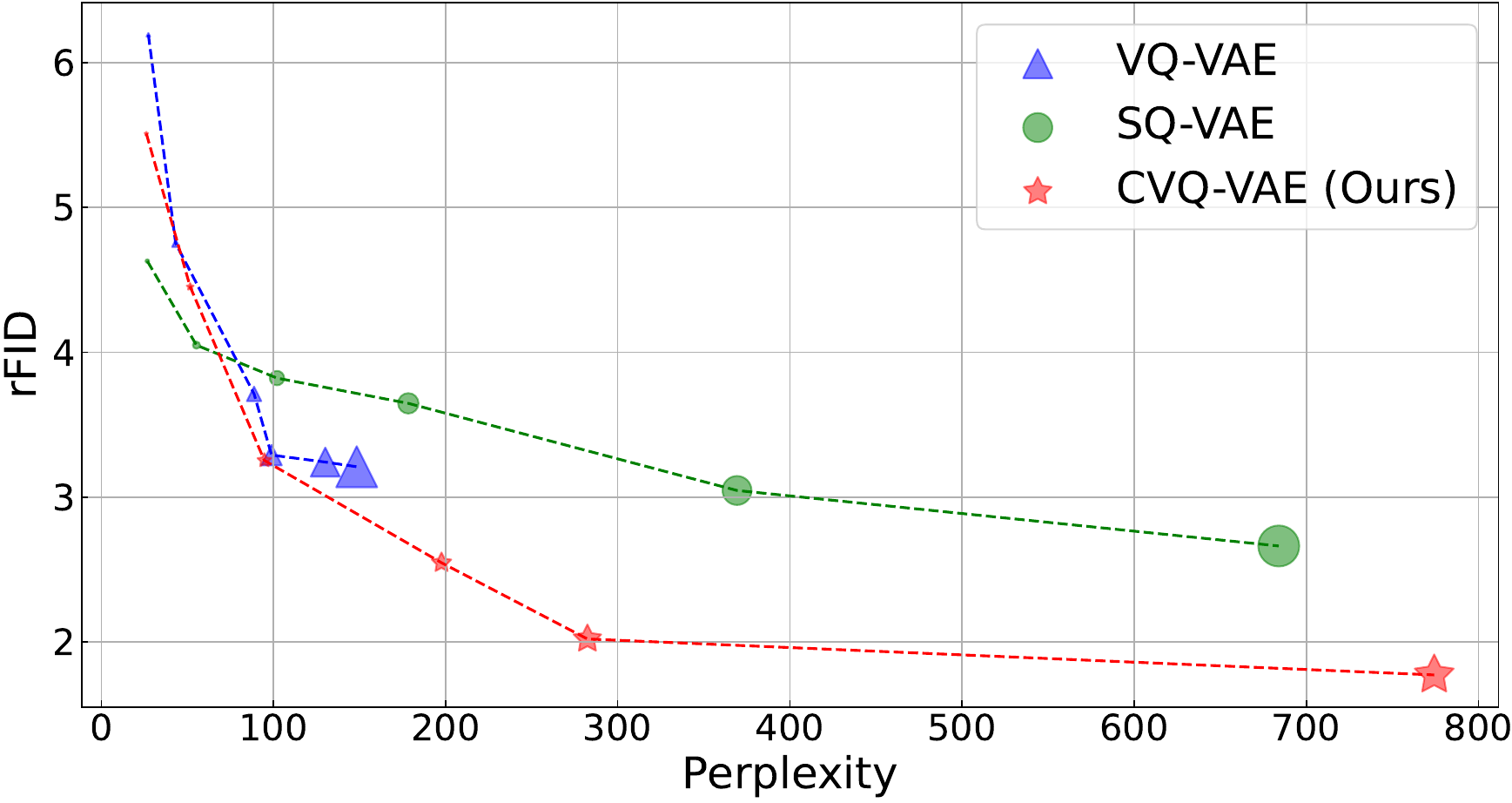}
        \includegraphics[width=\linewidth]{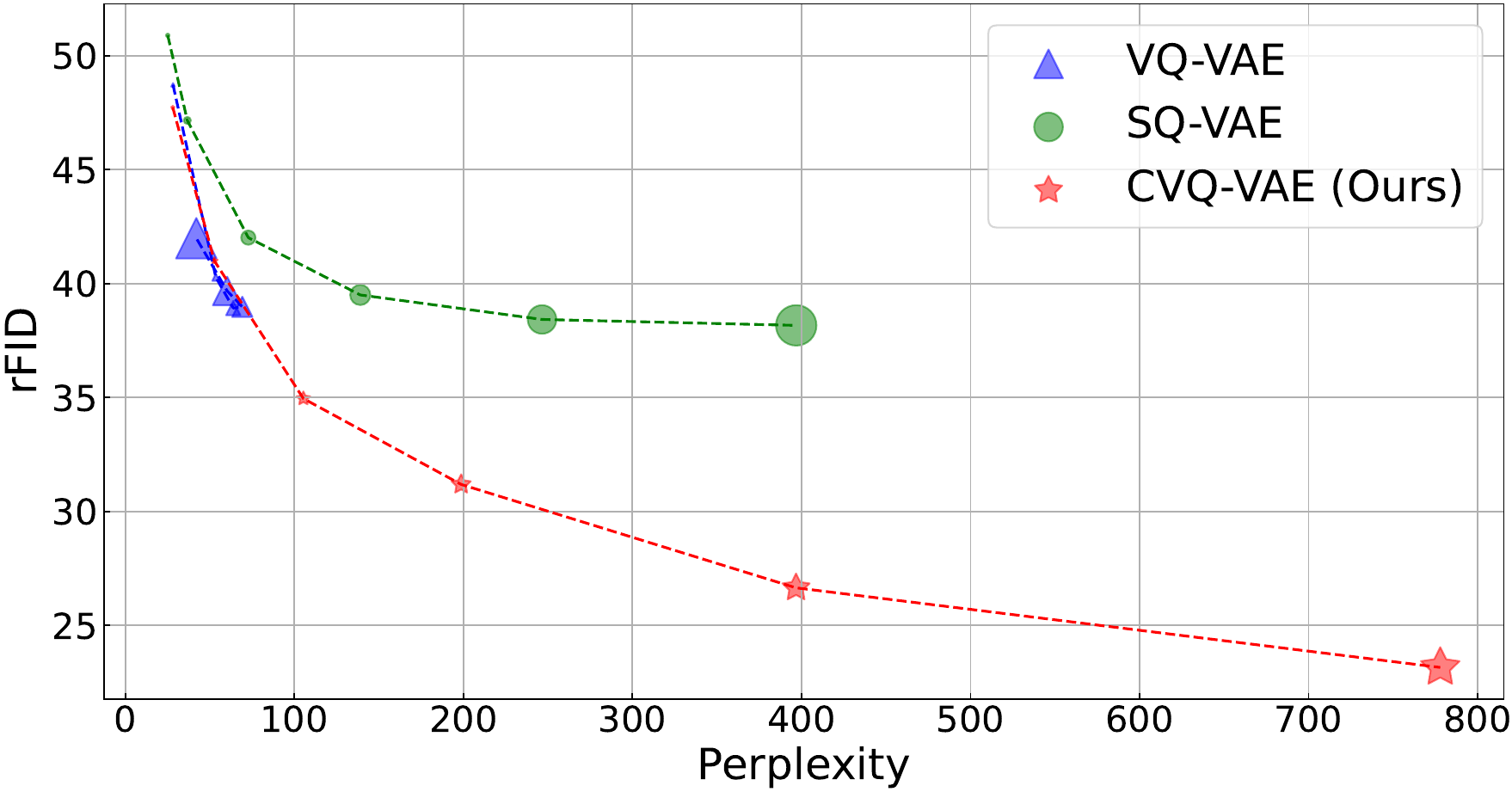}
    \end{center}
    }
    \end{minipage}
    }
    \hspace{0.5em}
    \subfloat[
    \textbf{Codebook dimensionality}. The blob's size refers to the dimensionality of codebook vectors \{4,8,16,32,64,128\}. The higher dimensionality does not ensure a better representation.
    \label{tab:rec_ab_codedim}
    ]{
    \centering
    \begin{minipage}{0.30\linewidth}
    {
    \begin{center}
        \includegraphics[width=\linewidth]{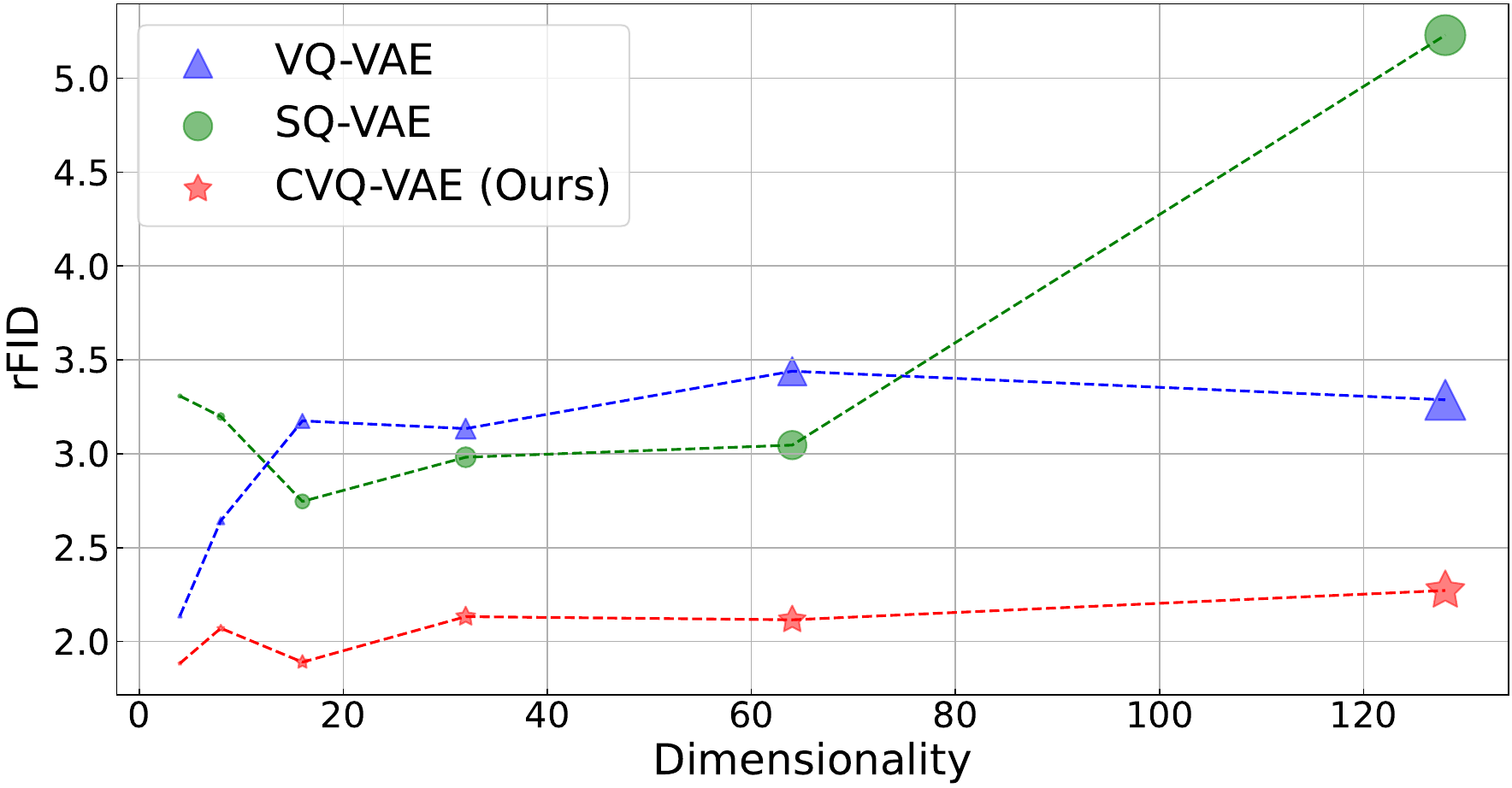}
        \includegraphics[width=\linewidth]{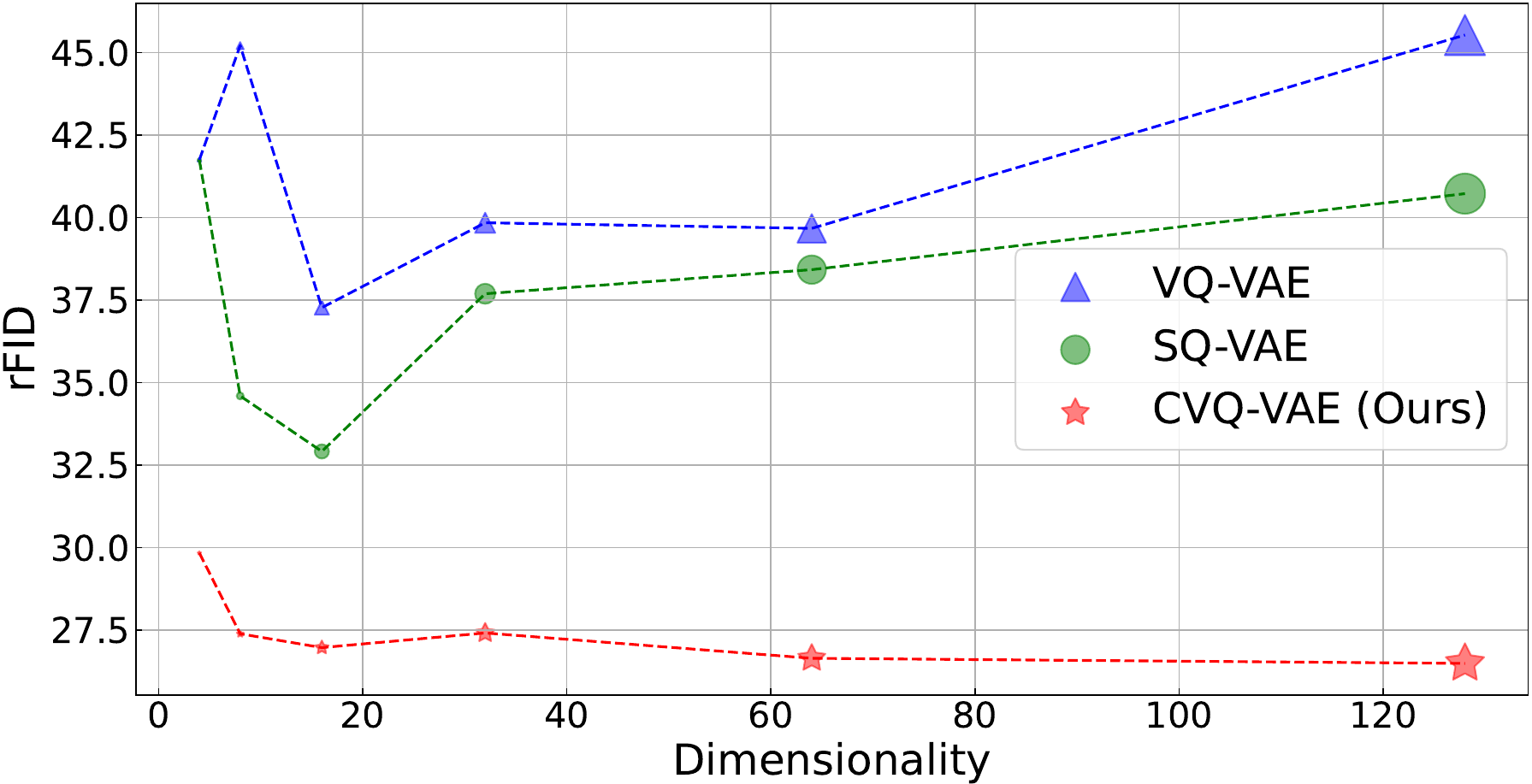}
    \end{center}
    }
    \end{minipage}
    }\hspace{0.5em}
    \subfloat[
    \textbf{Anchor sampling methods.}
    The choice of anchor sampling method has a significant impact on offline (one-time) feature initialization, while the online clustered method is robust for various samplings.%
    \label{tab:rec_ab_featm}
    ]{
    \centering
    \begin{minipage}{0.34\linewidth}
    {
    \begin{center}
        \renewcommand{\arraystretch}{1.15}
        \setlength\tabcolsep{3pt}
        \begin{tabular}{@{}l ccc  @{}}
        \toprule
        \multirow{2}{*}{\textbf{Method}} & \multirow{2}{*}{\textbf{Dataset}}& \multicolumn{2}{c}{rFID$\downarrow$}\\
        \cline{3-4}
        & & (offline) & (online) \\
        \midrule
        random & \multirow{4}{*}{MNIST} & 3.20 & 2.27 \\
        unique &&  2.84 & 2.24 \\
        probability &&  2.78 & \textbf{2.23} \\
        closest &&  \textbf{2.51} & 2.59 \\
        \midrule
        random & \multirow{4}{*}{CIFAR10}&  34.49 & 26.04 \\
        unique &&  36.99 & 26.02 \\
        probability &&  \textbf{31.10} & 26.62 \\
        closest &&  32.31 & \textbf{25.99} \\
        \bottomrule
        \end{tabular}
    \end{center}
    }
    \end{minipage}
    }
    \\
    \subfloat[
    \textbf{Codebook reinitialization methods.} In previous works~\cite{williams2020hierarchical,dhariwal2020jukebox}, each code entry is associated only with a single feature.
    \label{tab:rec_ab_reset}
    ]{
    \centering
    \begin{minipage}{\linewidth}
    {
    \begin{center}
        \renewcommand{\arraystretch}{1.2}
        \setlength\tabcolsep{5pt}
        \begin{tabular}{@{}l cccc cccc cccc @{}}
        \toprule
        \multirow{2}{*}{\textbf{Methods}}& \multicolumn{3}{c}{\textbf{MNIST}{ (28$\times$28)}} && \multicolumn{3}{c}{\textbf{CIFAR10}{ (32$\times$32)}} && \multicolumn{3}{c}{\textbf{FFHQ} (256$\times$256)} \\
        \cline{2-4}\cline{6-8}\cline{10-12}
         & SSIM $\uparrow$ & LPIPS $\downarrow$ & rFID$\downarrow$ && SSIM $\uparrow$ & LPIPS $\downarrow$ & rFID$\downarrow$&& SSIM $\uparrow$ & LPIPS $\downarrow$ & rFID$\downarrow$ \\
        \midrule
        near codevectors~\cite{williams2020hierarchical}  & 0.9790 & 0.0270 & 3.17 && 0.8553 & 0.2553 & 41.08 && 0.7282 & 0.1085 & 4.31 \\
        hard encoded features~\cite{dhariwal2020jukebox} & 0.9814 & 0.0243 & 2.25 && 0.8988 & 0.1978 & 29.16 && 0.7646 & 0.0870 & 3.91 \\
        running average (ours) & \textbf{0.9823} & \textbf{0.0236} & \textbf{2.23} && \textbf{0.8991} & \textbf{0.1897} & \textbf{26.62} && \textbf{0.8193} & \textbf{0.0603} & \textbf{2.94} \\
        \bottomrule
        \end{tabular}
    \end{center}
    }
    \end{minipage}
    }
    \caption{\textbf{Ablations} for \mname on image quantisation. We mainly train on MNIST and CIFAR10 training set, and evaluate on the validation set unless otherwise noted.}%
    \label{tab:rec_ab}
\end{table*}

\paragraph{Qualitative Results:}

The qualitative comparisons are presented in \cref{fig:rec}.
Our model achieves superior results even under challenging conditions.
Compared to the baseline model VQGAN~\cite{esser2021taming}, our \mname provides higher-quality reconstructed images that retain much more details.
In particular, VQGAN struggles with reconstructing abundant scene elements, as evidenced by the artifacts on the bowls.
In contrast, our \mname shows no such artifacts.
These fine-grained details are crucial for downstream generation-related tasks, such as generation, completion, and translation~\cite{zhengmovq}.

\subsection{Ablation Experiments}

We ran a number of ablations to analyse the effects of core factors in codebook learning.
Results are reported in \cref{tab:rule,tab:rec_ab,tab:app_rule,tab:app_rec_ab_featm}.

\paragraph{Core Factors.}

We evaluated core components in our redesigned online clustering quantiser in \cref{tab:rule}, which shows that the new quantiser considerably enhances the reconstruction quality.
We started by implementing the baseline configuration ($\mathbb{A}$) from VQ-VAE~\cite{van2017neural}.
Next, we explored different distance metrics, which are used to look up the closest entry for each encoded feature.
We found that using cosine similarity ($\mathbb{B}$) improved performance on some datasets, which is consistent with the findings in previous works such as ViT-VQGAN~\cite{yu2022vectorquantized}.
In configuration ($\mathbb{C}$), we reinitialized the unoptimized code entries with the selected anchors, but only in the first training batch, which we refer to as the \emph{offline} version.
This improved the usage of the codebook, resulting in slightly better gains.
Significantly, when we applied the proposed \emph{running average updates} across different training mini-batches in configuration ($\mathbb{D}$), the performance on all metrics in various datasets improved substantially.
This suggests that our proposed online clustering is significant for handling the changing encoded feature representation in deep networks.
Finally, we naturally introduced contrastive loss to each entry based on its similarity to features ($\mathbb{E}$), which further improved the results.

\paragraph{Codebook Size.}

VQ embeds the continuous features into a discrete space with a finite size, \ie $K$ codebook entries.
The codebook size has significant effects on traditional clustering. In \cref{tab:rec_ab_codesize}, we showed the performances of various quantisers with different codebook sizes.
Our \mname benefits greatly from a larger number of codebook entries, while SQ-VAE~\cite{takida2022sq} shows smaller improvements.
It is worth noting that \emph{not} all quantizers automatically benefit from a larger codebook size, such as VQ-VAE's performance on the CIFAR10 dataset shown in \cref{tab:rec_ab_codesize} (bottom).

\paragraph{Perplexity \emph{vs.} rFID.}

Recent concurrent studies~\cite{takida2022sq,gu2022rethinking,vuong2023vector} have explicitly promoted a large perplexity by optimizing a perplexity-related loss.
However, as illustrated in \cref{tab:rec_ab_codesize}, a larger perplexity does \emph{not} always guarantee a lower rFID\@.
This suggests that a uniform distribution of perplexity, represented by the highest score, may not be the optimal solution for the codebook.

\paragraph{Codebook Dimensionality.}

\Cref{tab:rec_ab_codedim} presents the results on various codebook dimensionalities.
Interestingly, the performance of the quantizers \emph{does not exhibit a straightforward positive correlation} with the number of codebook dimensionality.
In fact, some smaller codebook dimensionalities yield better results than larger ones, indicating that the choice of codebook dimensionality should be carefully considered depending on the specific application and dataset.
Based on this observation, a low-dimensional codebook can be employed to represent images and used in downstream tasks, as demonstrated in the latent diffusion model (LDM)~\cite{rombach2022high}.
The relevant downstream applications on generation can be found in \cref{sec:app}.

\paragraph{Anchor Sampling Methods.}

An evaluation of various \emph{anchor sampling methods} is reported in \cref{tab:rec_ab_featm}.
The results indicate that the \emph{offline} version with only one reinitialization is highly sensitive to the anchor sampling methods.
Interestingly, the random, unique, closest, and probabilistic random versions perform similarly for \emph{online} version, up to some random variations (rFID from 2.23 to 2.59 on MNIST, and from 25.99 to 26.62 on CIFAR10).
As discussed in \cref{sec:mth_online}, different anchor sampling methods have significant effects on traditional clustering~\cite{bradley1998refining,hamerly2002alternatives}.
However, our experiments demonstrate that the codebook reinitialization needs to consider the fact that \emph{the encoded features change along with the deep network is trained}.
The results highlight the effectiveness of our \emph{online} version with the running average updates, which is insensitive to the different instantiations.

\paragraph{Reinitialization Methods.}

Some latest works~\cite{williams2020hierarchical,dhariwal2020jukebox} also consider updating the unoptimized codevectors, called \emph{codebook reset}.
In \cref{tab:rec_ab_reset}, we compare these methods with VQ-VAE's architecture~\cite{van2017neural} under the same experimental setting, except for the different quantisers.
As discussed in \cref{sec:mth_online}, HVQ-VAE~\cite{williams2020hierarchical} resets the low usage codevectors using the high usage ones, which learns a narrow space codebook, resulting in limited improvement.
The hard encoded features presented in~\cite{dhariwal2020jukebox} achieve better results (3.17$\to$2.25, 41.08$\to$29.16, and 4.31$\to$3.91) than the HVQ-VAE~\cite{williams2020hierarchical} by adding noise signal to ensure the independent anchors for each codebook entry.
In contrast, our \mname calculates the running average updates, resulting in a significant improvement.
This further suggests that the online clustering centre along with the different training mini-batches is crucial for proper codebook reinitialization.

\section{Experiments: Applications}%
\label{sec:app}

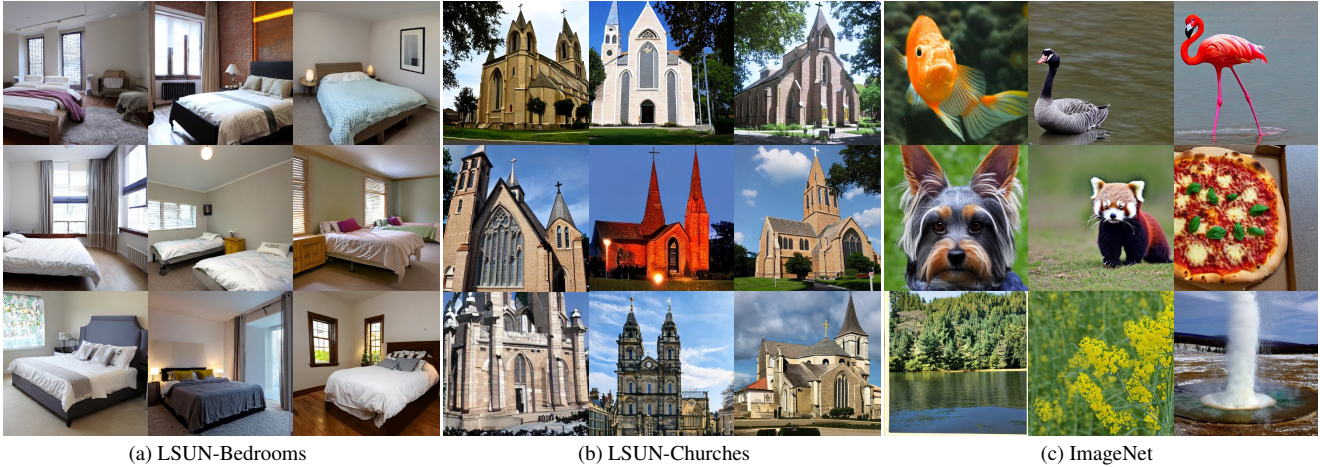
\begin{figure*}[tb!]
    \centering
    \includegraphics[width=\linewidth]{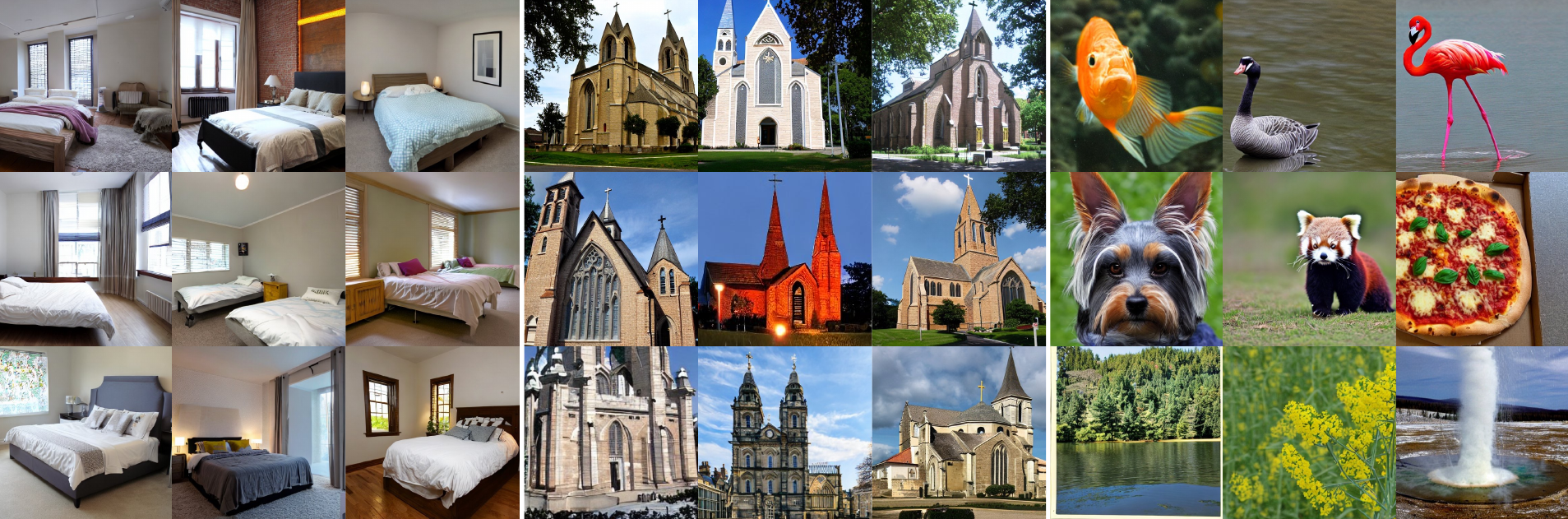}
    \begin{picture}(0,0)
    \put(-200,3){\footnotesize (a) LSUN-Bedrooms}
    \put(-30,3){\footnotesize (b) LSUN-Churches}
    \put(145,3){\footnotesize (c) ImageNet}
    \end{picture}
    \vspace{-0.2cm}
    \caption{\textbf{Unconditional} image generation on LSUN~\cite{yu2015lsun}, and \textbf{\emph{class}-conditional} image generation on Imagenet~\cite{deng2009imagenet}. Following the baseline LDM~\cite{rombach2022high}, our results are generated on 256$\times$256 resolution. Our training parameters are the same as in LDM, except for the different quantisers and 8$\times$ downsampling for the latent representations.}%
    \label{fig:ge}
\end{figure*}

Except for data compression, our \mname can also be easily applied to downstream tasks, such as generation and completion.
Following existing works~\cite{esser2021taming,rombach2022high,zhengmovq}, we conduct a simple experiment to verify the effectiveness of the proposed quantisers.
Although this simple yet effective quantiser can be applied to more applications, it is beyond the main scope of this paper.

\begin{table}[tb!]
    \centering
    \setlength\tabcolsep{5pt}
        \begin{tabular}{@{}l cc @{}}
        \toprule
         \multirow{2}{*}{\textbf{Methods}} &  \multicolumn{2}{c}{FID$\downarrow$}\\
         \cline{2-3}
         & Churches & Bedrooms \\
         \midrule
         StyleGAN~\cite{karras2019style} & 4.21 & 2.35\\
         DDPM~\cite{ho2020denoising} & 7.89 & 4.90 \\
         ImageBART~\cite{esser2021imagebart} & 7.32 & 5.51 \\
         Projected-GAN~\cite{sauer2021projected} & \textbf{1.59} & \textbf{1.52} \\
         \midrule
         LDM~\cite{rombach2022high}-8$^*$ & 4.02 & - \\
         LDM~\cite{rombach2022high}-4 & - & 2.95 \\
         \midrule
         LDM~\cite{rombach2022high}-8 (reproduced) & 4.15 & 3.57 \\
         \mname-LDM~\cite{rombach2022high}-8 & 3.86 & 3.02 \\
         \bottomrule
    \end{tabular}
    \caption{\textbf{Quantitative comparisons on unconditional image generation.} The better quantiser can improve the generation quality without modifying the training settings in the second stage. $^*$: trained in $KL$-regularized latent space, instead of the VQ discrete space.}%
    \label{tab:ge}
\end{table}

\begin{table}[tb!]
    \centering
    \setlength\tabcolsep{5pt}
    \begin{tabular}{@{}l ccc cc@{}}
    \toprule
     \multirow{2}{*}{\textbf{Model}} & \multicolumn{2}{c}{\textbf{FFHQ}} && \multicolumn{2}{c}{\textbf{ImageNet}} \\
     \cline{2-3}\cline{5-6}
     & Steps & FID$\downarrow$ &&  Steps & FID$\downarrow$ \\
    \midrule
     RQVAE [22]$_{\text{\scriptsize{CVPR'2022}}}$  & 256 & 10.38 && 1024 & 7.55 \\
     MoVQ [44]$_{\text{\scriptsize{NeurIPS'2022}}}$   & 1024 & 8.52 && 1024 & 7.13 \\
     SQ-VAE [33]$_{\text{\scriptsize{ICML'2022}}}$   & 200 & 5.17 && 250 & 9.31 \\
     LDM-4 [31]$_{\text{\scriptsize{CVPR'2022}}}$  & 200 & 4.98 && 250 & 10.56 \\
     \textbf{CVQ-VAE} (ours) & 200 & \textbf{4.46} && 250 &  \textbf{6.87} \\
    \bottomrule
    \end{tabular}
    \caption{Quantitative results for unconditional generation on FFHQ and \emph{class}-conditional generation on ImageNet.}
    \label{tab:ge-c}
\end{table}

\paragraph{Implementation Details.}

We made minor modifications to the baseline LDM~\cite{rombach2022high} system when adapting it with our quantiser for the downstream tasks.
We first replace the original quantiser from VQGAN~\cite{esser2021taming} with our proposed \mname's quantiser.
Then, we trained the models on LSUN~\cite{yu2015lsun} and ImageNet~\cite{deng2009imagenet} for generation (8$\times$ downsampling).
Following the setting in LDM~\cite{rombach2022high}, we set the sampling step as 200 during the inference. 

\subsection{Unconditional Generation}

\Cref{tab:ge,tab:ge-c} compares our proposed \mname to the state-of-the-art methods on LSUN and ImageNet datasets for unconditional and \emph{class}-conditional image generation.
The results show that our model consistently improves the generated image quality under the sample compression ratio, as in the reconstruction.
This confirms the advantages of using a better codebook for downstream tasks.
Our \mname also outperforms the LDM-8$^*$ that is trained with $KL$-regularized latent space, indicating that exploring a better discrete codebook is worth pursuing for unsupervised representation learning.
Our \mname also achieves comparable results to LDM-4 (3.02 \emph{vs.} 2.95), whereas the LDM-4 uses a 4$\times$ higher resolution representation, requiring more computational costs.

Example results are presented in \cref{fig:ge}.
As we can see, even with 8$\times$ downsampling, the proposed \mname is still able to generate reasonable structures for these complicated scenes with various instances.
Although there are artifacts on windows in the two scenarios, the other high-frequency details are realistic, such as the sheet on the bed.


\section{Conclusion and Limitation}

We have introduced \mname, a novel codebook reinitialization method that tackles the \emph{codebook collapse} issue by assigning the online clustered anchors to unoptimized code entries.
Our proposed quantiser is a simple yet effective solution that can be integrated into many existing architectures for representation learning.
Experimental results show that our \mname significantly outperforms the state-of-the-art VQ models on image modeling, yet without increasing computational cost and latent size.
We hope this new plug-and-play quantiser will become an important component of future vector methods that use VQ in their learned architecture.

\paragraph{Ethics.}

We use the MNIST, Fashion-MNIST, CIFAR10, LSUN, and ImageNet datasets in a manner compatible with their terms.
While some of these images contain personal information (\eg, faces) collected without consent, algorithms in this research do not extract biometric information.
For further details on ethics, data protection, and copyright please see \url{https://www.robots.ox.ac.uk/~vedaldi/research/union/ethics.html}.

\paragraph{Acknowledgements.}

This research is supported by ERC-CoG UNION 101001212.

{\small \bibliographystyle{ieee_fullname} \bibliography{egbib}}

\input{supplementary}

\end{document}

%% file: supplementary.tex
\appendix\onecolumn
\renewcommand{\theequation}{\thesection.\arabic{equation}}
\setcounter{equation}{0}
\renewcommand{\thefigure}{\thesection.\arabic{figure}}
\setcounter{figure}{0}
\renewcommand{\thetable}{\thesection.\arabic{table}}
\setcounter{table}{0}
\newpage

\begin{center}
\textbf{\Large Online Clustered Codebook} \\[5pt]
\end{center}


\section{Experiment Details}

For data compression, we first demonstrate our method on small datasets with the officially released VQ-VAE~\cite{van2017neural} implementation~\footnote{\href{https://github.com/deepmind/sonnet/blob/v2/sonnet/src/nets/vqvae.py}{https://github.com/deepmind/sonnet/blob/v2/sonnet/src/nets/vqvae.py}}\footnote{\href{https://github.com/deepmind/sonnet/blob/v1/sonnet/examples/vqvae_example.ipynb}{https://github.com/deepmind/sonnet/blob/v1/sonnet/examples/vqvae\_
example.ipynb}}, and we then verify the generality of our quantiser on large datasets using the officially released VQ-GAN~\cite{esser2021taming} architecture~\footnote{\href{https://github.com/CompVis/taming-transformers}{https://github.com/CompVis/taming-transformers}}.
For image generation application, we apply our \mname's quantiser on LSUN dataset using the officially released LDM~\cite{rombach2022high} code\footnote{\href{https://github.com/CompVis/latent-diffusion}{https://github.com/CompVis/latent-diffusion}}.

For the small datasets (MNIST, CIFAR10, and Fashion MNIST), we use the submitted \textcolor[RGB]{255,0,0}{code} to train the model.
The training hyperparameters match the original VQ-VAE settings, and we train all models with batch size 1,024 across 4$\times$ NVIDIA GeForce GTX TITAN X (12GB per GPU) with 500 epochs (2-3 hours).

For the high resolution datasets (FFHQ and ImageNet), we just replace the original quantiser in VQGAN with our \mname quantiser.
The training hyperparameters also follow the original settings, and we train all models with batch size 64 across 4$\times$ NVIDIA RTX A6000 (48GB per GPU) with 4 days on FFHQ and 8 days on ImageNet until converge.

For the generation (LSUN bedrooms and Churches), we use the lSUN-beds256 config file for default setting with two modifications: 1) we also replace the VQGAN's quantiser with our \mname quantiser; 2) we reduce the images' resolution for faster training with 8$\times$ downsampling. For stage \textbf{a)} codebook learning, two models are trained with batch size 32 across 2$\times$ NVIDIA RTX A4000 (48GB per GPU) with 5 days. Then, for stage \textbf{b)} latent diffusion model with $32\times32\times4$ resolution, we train the models with batch size 128 across 2$\times$ NVIDIA RTX A4000 (48GB per GPU) with 7 days. During the inference, we follow the default settings to sample the images with 200 steps.

\section{Quantitative Results}

\begin{table}[htb!]
    \centering
    \renewcommand{\arraystretch}{1.0}
    \setlength\tabcolsep{6pt}
    \begin{tabular}{@{}l c ccccc @{}}
    \toprule
    \textbf{Method} & \textbf{Dataset} & $\ell_1 \text{loss}\downarrow$ & SSIM $\uparrow$ & PSNR $\uparrow$ & LPIPS $\downarrow$ & rFID $\downarrow$ \\
    \midrule
    VQ-VAE~\cite{van2017neural} & \multirow{4}{*}{MNIST}& 0.0207 & 0.9777 & 26.48 & 0.0282 & 3.43 \\
    HVQ-VAE~\cite{williams2020hierarchical} & & 0.0202 & 0.9790 & 26.90 & 0.0270 & 3.17 \\
    SQ-VAE~\cite{takida2022sq} & & 0.0197 & 0.9819 & 27.49 & 0.0256 & 3.05 \\
    \cdashline{1-5}
    \textbf{\mname} & & \textbf{0.0180} &  \textbf{0.9833} & \textbf{27.87} &\textbf{0.0222} & \textbf{1.80} \\
    \midrule
    VQ-VAE~\cite{van2017neural} & \multirow{4}{*}{CIFAR10}& 0.0527 & 0.8595 & 23.32 & 0.2504 & 39.67 \\
    HVQ-VAE~\cite{williams2020hierarchical} & & 0.0533 &  0.8553 & 23.22 & 0.2553 & 41.08 \\
    SQ-VAE~\cite{takida2022sq} & & 0.0482 & 0.8779 & 24.07 & 0.2333 & 37.92 \\
    \cdashline{1-5}
    \textbf{\mname} & & \textbf{0.0448} & \textbf{0.8978} & \textbf{24.72} & \textbf{0.1883} & \textbf{24.73} \\
    \bottomrule
    \end{tabular}
    \caption{\textbf{Reconstruction results} on the validation sets of MNIST (10,000 images) and CIFAR10 (10,000 images). }
    \label{tab:app_rec_quant}
\end{table}

\Cref{tab:app_rec_quant} provides a comparison of our results to state-of-the-art quantisers under the same training settings, except for the different quantisers, on the small datasets.
This is an extension of \cref{tab:rec_quant} in the main paper.
All images are normalised to the range [0,1] for quantitative evaluation.
See the code for more details.
While the proposed \mname achieve relative small improvements on traditional pixel-level $\ell_1$ loss, peak signal-to-noise ration (PSNR), and patch-level structure similarity index (SSIM), it significantly improves the feature-level LPIPS and dataset-level rFID, suggesting that our \mname is more capable of reconstructing the content closer to the dataset distribution.

\begin{table}[htb!]
    \centering
    \renewcommand{\arraystretch}{1.0}
    \setlength\tabcolsep{6pt}
    \begin{tabular}{@{}l c ccccccc @{}}
    \toprule
         \textbf{Method} &  \textbf{Dataset} & $\mathcal{S}$ $\downarrow$ & $\mathcal{K}$ $\downarrow$ & Usage $\uparrow$ & PSNR $\uparrow$ & SSIM $\uparrow$ & LPIPS $\downarrow$ & rFID $\downarrow$ \\
    \midrule
    VQGAN~\cite{esser2021taming} & \multirow{6}{*}{FFHQ}& 16$^2$& 1024& $42\%$ & 22.24 & 0.6641 & 0.1175 & 4.42\\
    ViT-VQGAN~\cite{yu2022vectorquantized} & & 32$^2$& 8192& -- & -- & -- & -- & 3.13\\
    RQ-VAE~\cite{lee2022autoregressive} & & 16$^2\times$4 & 2048 & -- & 22.99 & 0.6700 & 0.1302 & 3.88 \\
    MoVQ~\cite{zhengmovq} & & 16$^2\times$4 & 1024 & $56\%$& 26.72 & 0.8212 & 0.0585 & 2.26 \\
    SeQ-GAN~\cite{gu2022rethinking} & & 16$^2$& 1024 & $100\%$& -- & -- & -- & 3.12 \\
    \cdashline{1-9}
    \textbf{\mname} (ours) & & 16$^2$& 1024 & $100\%$ & 26.82 & 0.8313 & 0.0608 & 2.80 \\
    \textbf{\mname} (ours) & & 16$^2\times$4 & 1024 & $100\%$ & \textbf{26.87} & \textbf{0.8398} & \textbf{0.0533} & \textbf{2.03}\\
    \midrule
    VQGAN~\cite{esser2021taming} & \multirow{6}{*}{ImageNet}& 16$^2$& 1024 & $44\%$ & 19.07 & 0.5183 & 0.2011 & 7.94 \\
    ViT-VQGAN~\cite{yu2022vectorquantized} & & 32$^2$& 8192 & $96\%$& -- & -- & -- & 1.28 \\
    RQ-VAE~\cite{lee2022autoregressive} & & 8$^2\times$16& 16384 & - & -- & -- & -- & 1.83\\
    MoVQ~\cite{zhengmovq} & & 16$^2\times$4& 1024 & $63\%$& 22.42 & 0.6731 & 0.1132 & \textbf{1.12} \\
    SeQ-GAN~\cite{gu2022rethinking} & & 16$^2$ & 1024 & $100\%$ & -- & -- & -- & 1.99\\
    \cdashline{1-9}
    \textbf{\mname} (ours) & & 16$^2$& 1024 & $100\%$ & 21.95 & 0.6612 & 0.1340 & 1.57 \\
    \textbf{\mname} (ours) & & 16$^2\times$4 & 1024 & $100\%$ & \textbf{23.37} & \textbf{0.7115} & \textbf{0.1099}& 1.20 \\
    \bottomrule
    \end{tabular}
    \caption{\textbf{Reconstruction results} on validation sets of ImageNet (50,000 images) and FFHQ (10,000 images). $\mathcal{S}$ denotes the latent size of encoded features, and $\mathcal{K}$ is the number of codevectors in the codebook. }
    \label{tab:app_rec_sota}
\end{table}

We further compare our \mname to the state-of-the-art methods in data compression in \cref{tab:app_rec_sota}.
This is an extension of \cref{tab:rec_sota} in the main paper.
Here, we add the pixel-level PSNR, patch-level SSIM and feature-level LPIPS.
For FFHQ dataset, our \mname model outperforms baseline variants of previous state-of-the-art models.
As for ImageNet dataset, while our 4$\times$ channels setting does not achieve the better rFID than the latest MoVQ model, the other instantiations (PSNR, SSIM and LPIPS) significantly outperform existing state-of-the-art models.

\begin{table}[htb!]
    \centering
    \renewcommand{\arraystretch}{1.0}
    \setlength\tabcolsep{3pt}
    \begin{tabular}{@{}ll cccc cccc cccc@{}}
    \toprule
         &  \multirow{2}{*}{\textbf{Method}} & \multicolumn{3}{c}{\textbf{MNIST} (28$\times$28)} && \multicolumn{3}{c}{\textbf{CFAIR10} (32$\times$32)} && \multicolumn{3}{c}{\textbf{Fashion MNIST} (28$\times$28)}\\
         \cline{3-5}\cline{7-9}\cline{11-13}
         &  & $\ell_1 \downarrow$ & PSNR $\uparrow$ & rFID $\downarrow$ && $\ell_1 \downarrow$ & PSNR $\uparrow$ & rFID $\downarrow$ && $\ell_1 \downarrow$ & PSNR $\uparrow$ & rFID $\downarrow$\\
    \midrule
    $\mathbb{A}$ & Baseline VQ-VAE~\cite{van2017neural}$_{\text{\scriptsize{NeurIPS'2017}}}$ & 0.0207 & 26.48 & 3.43 && 0.0527 & 23.32 & 39.67 && 0.0377 & 23.93 & 12.73 \\
    \cdashline{1-13}
    $\mathbb{B}$ & + Cosine distance & 0.0200 & 26.77 & 3.06 && 0.0509 & 23.66 & 35.14 && 0.0378 & 24.01 & 11.40 \\
    $\mathbb{C}$ & + Anchor initialization (offline) &  0.0192 & 27.24 & 2.78 && 0.0481 & 24.16 & 31.10 && 0.0373 & 24.04 & 11.92 \\
    $\mathbb{D}$ & + Anchor initialization (online) & 0.0186 & 27.58 & 2.23 && \textbf{0.0445} & \textbf{24.79} & 26.62  && 0.0349 & \textbf{24.69} & 9.27 \\
    $\mathbb{E}$ & + Contrastive loss &  \textbf{0.0180} & \textbf{27.87} & \textbf{1.80} && 0.0448 & 24.72 & \textbf{24.73} && \textbf{0.0344} & 24.66 & \textbf{8.85}\\
    \bottomrule
    \end{tabular}
    \vspace{-0.1cm}
    \caption{\textbf{Results on various settings.} We add pixel-level $\ell_1$ and PSNR metrics.}
    \vspace{-0.3cm}
    \label{tab:app_rule}
\end{table}

\begin{table}[htb!]
    \centering
    \renewcommand{\arraystretch}{1.15}
        \setlength\tabcolsep{3pt}
        \begin{tabular}{@{}l c ccccc c ccccc  @{}}
        \toprule
        \multirow{2}{*}{\textbf{Method}} & \multirow{2}{*}{\textbf{Dataset}}& \multicolumn{5}{c}{Offline} && \multicolumn{5}{c}{Online} \\
        \cline{3-7}\cline{9-13}
        & & $\ell_1 \text{loss}\downarrow$ & SSIM $\uparrow$ & PSNR $\uparrow$ & LPIPS $\downarrow$ & rFID $\downarrow$ && $\ell_1 \text{loss}\downarrow$ & SSIM $\uparrow$ & PSNR $\uparrow$ & LPIPS $\downarrow$ & rFID $\downarrow$ \\
        \midrule
        random & \multirow{4}{*}{MNIST} & 0.0195 & 0.9802 & 27.11 & 0.0262 & 3.20 && \textbf{0.0185} & \textbf{0.9823} & \textbf{27.58} & \textbf{0.0236} & 2.27\\
        unique && 0.0191 & 0.9811 & 27.25 & 0.0255 & 2.84 && 0.0186 & 0.9820 & 27.51 & 0.0237 & 2.24 \\
        probability && 0.0192 & 0.9810 & 27.24 & 0.0253 & 2.78 && 0.0186 & \textbf{0.9823} & \textbf{27.58} & \textbf{0.0236} & \textbf{2.23}\\
        closest && \textbf{0.0186} & \textbf{0.9823} & \textbf{27.59} & \textbf{0.0242} & \textbf{2.51} && 0.0187 & 0.9819 & 27.49 & 0.0244 & 2.59 \\
        \midrule
        random & \multirow{4}{*}{CIFAR10} & 0.0494 & 0.8755 & 23.91 & 0.2256 & 34.49 && 0.0440 & \textbf{0.9010} & 24.88 & 0.1881 & 26.04 \\
        unique && 0.0507 & 0.8705 & 23.15 & 0.2346 & 36.99 && \textbf{0.0439} & 0.9007 & \textbf{24.91} & \textbf{0.1877} & 26.03 \\
        probability && \textbf{0.0481} & \textbf{0.8829} & \textbf{24.16} & \textbf{0.2131}& \textbf{31.10} && 0.0445 & 0.8991 & 24.79 & 0.1898 & 26.62 \\
        closest && 0.0487 & 0.8804 & 24.06 & 0.2156 & 32.31 && 0.0444 & 0.8994 & 24.83 & 0.1900 & \textbf{25.99} \\
        \bottomrule
        \end{tabular}
    \caption{\textbf{Anchor sampling methods.} The choice of anchor sampling method has a significant impact on
    offline (one-time) feature initialization, while the online clustered method is robust for various samplings.}
    \label{tab:app_rec_ab_featm}
\end{table}

\Cref{tab:app_rule,tab:app_rec_ab_featm} are the extension of \cref{tab:rule,tab:rec_ab_featm} in the main paper, respectively.
Even reported with the different metrics, The conclusions are still the same.
For instance, the offline version is significantly affected by different anchor sampling methods, but the online version is not sensitive to various anchor sampling methods.
The online version holds very close performance with these anchor sampling methods.




%% file: main.bbl
\begin{thebibliography}{10}\itemsep=-1pt

\bibitem{agustsson2017soft}
Eirikur Agustsson, Fabian Mentzer, Michael Tschannen, Lukas Cavigelli, Radu
  Timofte, Luca Benini, and Luc~V Gool.
\newblock Soft-to-hard vector quantization for end-to-end learning compressible
  representations.
\newblock In {\em Proceedings of the 31st International Conference on Neural
  Information Processing Systems (NeurIPS)}, volume~30, 2017.

\bibitem{arthur2007k}
D ARTHUR.
\newblock k-means++: the advantages of careful seeding.
\newblock In {\em Proceedings of the eighteenth annual ACM-SIAM symposium on
  Discrete algorithms, New Orleans, Louisiana, 2007}, pages 1027--1035. Society
  for Industrial and Applied Mathematics, 2007.

\bibitem{baobeit}
Hangbo Bao, Li Dong, Songhao Piao, and Furu Wei.
\newblock Beit: Bert pre-training of image transformers.
\newblock In {\em Proceedings of the International Conference on Learning
  Representations (ICLR)}, 2022.

\bibitem{bradley1998refining}
Paul~S Bradley and Usama~M Fayyad.
\newblock Refining initial points for k-means clustering.
\newblock In {\em ICML}, volume~98, pages 91--99. Citeseer, 1998.

\bibitem{celebi2013comparative}
M~Emre Celebi, Hassan~A Kingravi, and Patricio~A Vela.
\newblock A comparative study of efficient initialization methods for the
  k-means clustering algorithm.
\newblock {\em Expert systems with applications}, 40(1):200--210, 2013.

\bibitem{chang2022maskgit}
Huiwen Chang, Han Zhang, Lu Jiang, Ce Liu, and William~T. Freeman.
\newblock Maskgit: Masked generative image transformer.
\newblock In {\em The IEEE Conference on Computer Vision and Pattern
  Recognition (CVPR)}, June 2022.

\bibitem{deng2009imagenet}
Jia Deng, Wei Dong, Richard Socher, Li-Jia Li, Kai Li, and Li Fei-Fei.
\newblock Imagenet: A large-scale hierarchical image database.
\newblock In {\em 2009 IEEE conference on computer vision and pattern
  recognition (CVPR)}, pages 248--255. Ieee, 2009.

\bibitem{dhariwal2020jukebox}
Prafulla Dhariwal, Heewoo Jun, Christine Payne, Jong~Wook Kim, Alec Radford,
  and Ilya Sutskever.
\newblock Jukebox: A generative model for music.
\newblock {\em arXiv preprint arXiv:2005.00341}, 2020.

\bibitem{ding2021cogview}
Ming Ding, Zhuoyi Yang, Wenyi Hong, Wendi Zheng, Chang Zhou, Da Yin, Junyang
  Lin, Xu Zou, Zhou Shao, Hongxia Yang, et~al.
\newblock Cogview: Mastering text-to-image generation via transformers.
\newblock {\em Advances in Neural Information Processing Systems (NeurIPS)},
  34, 2021.

\bibitem{esser2021imagebart}
Patrick Esser, Robin Rombach, Andreas Blattmann, and Bjorn Ommer.
\newblock Imagebart: Bidirectional context with multinomial diffusion for
  autoregressive image synthesis.
\newblock In {\em Advances in Neural Information Processing Systems (NeurIPS)},
  volume~34, 2021.

\bibitem{esser2021taming}
Patrick Esser, Robin Rombach, and Bjorn Ommer.
\newblock Taming transformers for high-resolution image synthesis.
\newblock In {\em Proceedings of the IEEE/CVF conference on computer vision and
  pattern recognition (CVPR)}, pages 12873--12883, 2021.

\bibitem{1162229}
R. Gray.
\newblock Vector quantization.
\newblock {\em IEEE ASSP Magazine}, 1(2):4--29, 1984.

\bibitem{gu2022rethinking}
Yuchao Gu, Xintao Wang, Yixiao Ge, Ying Shan, Xiaohu Qie, and Mike~Zheng Shou.
\newblock Rethinking the objectives of vector-quantized tokenizers for image
  synthesis.
\newblock {\em arXiv preprint arXiv:2212.03185}, 2022.

\bibitem{hamerly2002alternatives}
Greg Hamerly and Charles Elkan.
\newblock Alternatives to the k-means algorithm that find better clusterings.
\newblock In {\em Proceedings of the eleventh international conference on
  Information and knowledge management}, pages 600--607, 2002.

\bibitem{heusel2017gans}
Martin Heusel, Hubert Ramsauer, Thomas Unterthiner, Bernhard Nessler, and Sepp
  Hochreiter.
\newblock Gans trained by a two time-scale update rule converge to a local nash
  equilibrium.
\newblock In {\em Proceedings of the 31st International Conference on Neural
  Information Processing Systems (NeurIPS)}, pages 6626--6637, 2017.

\bibitem{ho2020denoising}
Jonathan Ho, Ajay Jain, and Pieter Abbeel.
\newblock Denoising diffusion probabilistic models.
\newblock {\em Advances in Neural Information Processing Systems (NeurIPS)},
  33:6840--6851, 2020.

\bibitem{hu2022global}
Minghui Hu, Yujie Wang, Tat-Jen Cham, Jianfei Yang, and Ponnuthurai~N
  Suganthan.
\newblock Global context with discrete diffusion in vector quantised modelling
  for image generation.
\newblock In {\em Proceedings of the IEEE/CVF Conference on Computer Vision and
  Pattern Recognition (CVPR)}, pages 11502--11511, 2022.

\bibitem{hu2022unified}
Minghui Hu, Chuanxia Zheng, Heliang Zheng, Tat-Jen Cham, Chaoyue Wang, Zuopeng
  Yang, Dacheng Tao, and Ponnuthurai~N Suganthan.
\newblock Unified discrete diffusion for simultaneous vision-language
  generation.
\newblock {\em arXiv preprint arXiv:2211.14842}, 2022.

\bibitem{karras2019style}
Tero Karras, Samuli Laine, and Timo Aila.
\newblock A style-based generator architecture for generative adversarial
  networks.
\newblock In {\em Proceedings of the IEEE/CVF conference on computer vision and
  pattern recognition}, pages 4401--4410, 2019.

\bibitem{krizhevsky2009learning}
Alex Krizhevsky, Geoffrey Hinton, et~al.
\newblock Learning multiple layers of features from tiny images.
\newblock 2009.

\bibitem{lecun1998gradient}
Yann LeCun, L{\'e}on Bottou, Yoshua Bengio, and Patrick Haffner.
\newblock Gradient-based learning applied to document recognition.
\newblock {\em Proceedings of the IEEE}, 86(11):2278--2324, 1998.

\bibitem{lee2022autoregressive}
Doyup Lee, Chiheon Kim, Saehoon Kim, Minsu Cho, and Wook-Shin Han.
\newblock Autoregressive image generation using residual quantization.
\newblock In {\em Proceedings of the IEEE/CVF Conference on Computer Vision and
  Pattern Recognition (CVPR)}, pages 11523--11532, 2022.

\bibitem{li2022unimo}
Wei Li, Can Gao, Guocheng Niu, Xinyan Xiao, Hao Liu, Jiachen Liu, Hua Wu, and
  Haifeng Wang.
\newblock Unimo-2: End-to-end unified vision-language grounded learning.
\newblock In {\em Findings of the Association for Computational Linguistics:
  ACL 2022}, pages 3187--3201, 2022.

\bibitem{liu2022cross}
Alex Liu, SouYoung Jin, Cheng-I Lai, Andrew Rouditchenko, Aude Oliva, and James
  Glass.
\newblock Cross-modal discrete representation learning.
\newblock In {\em Proceedings of the 60th Annual Meeting of the Association for
  Computational Linguistics (Volume 1: Long Papers)}, pages 3013--3035, 2022.

\bibitem{lloyd1982least}
Stuart Lloyd.
\newblock Least squares quantization in pcm.
\newblock {\em IEEE transactions on information theory}, 28(2):129--137, 1982.

\bibitem{maodiscrete}
Chengzhi Mao, Lu Jiang, Mostafa Dehghani, Carl Vondrick, Rahul Sukthankar, and
  Irfan Essa.
\newblock Discrete representations strengthen vision transformer robustness.
\newblock In {\em Proceedings of the International Conference on Learning
  Representations (ICLR)}, 2022.

\bibitem{mittal2022autosdf}
Paritosh Mittal, Yen-Chi Cheng, Maneesh Singh, and Shubham Tulsiani.
\newblock Autosdf: Shape priors for 3d completion, reconstruction and
  generation.
\newblock In {\em Proceedings of the IEEE/CVF Conference on Computer Vision and
  Pattern Recognition(CVPR)}, pages 306--315, 2022.

\bibitem{rakhimov2021latent}
Ruslan Rakhimov, Denis Volkhonskiy, Alexey Artemov, Denis Zorin, and Evgeny
  Burnaev.
\newblock Latent video transformer.
\newblock In {\em 16th International Joint Conference on Computer Vision,
  Imaging and Computer Graphics Theory and Applications, VISIGRAPP 2021}, pages
  101--112. SciTePress, 2021.

\bibitem{ramesh2022hierarchical}
Aditya Ramesh, Prafulla Dhariwal, Alex Nichol, Casey Chu, and Mark Chen.
\newblock Hierarchical text-conditional image generation with clip latents.
\newblock {\em arXiv preprint arXiv:2204.06125}, 2022.

\bibitem{ramesh2021zero}
Aditya Ramesh, Mikhail Pavlov, Gabriel Goh, Scott Gray, Chelsea Voss, Alec
  Radford, Mark Chen, and Ilya Sutskever.
\newblock Zero-shot text-to-image generation.
\newblock In {\em International Conference on Machine Learning (ICML)}, pages
  8821--8831. PMLR, 2021.

\bibitem{razavi2019generating}
Ali Razavi, Aaron Van~den Oord, and Oriol Vinyals.
\newblock Generating diverse high-fidelity images with vq-vae-2.
\newblock In {\em Advances in Neural Information Processing Systems (NeurIPS)},
  volume~32, 2019.

\bibitem{rombach2022high}
Robin Rombach, Andreas Blattmann, Dominik Lorenz, Patrick Esser, and Bj{\"o}rn
  Ommer.
\newblock High-resolution image synthesis with latent diffusion models.
\newblock In {\em Proceedings of the IEEE/CVF Conference on Computer Vision and
  Pattern Recognition(CVPR)}, pages 10684--10695, 2022.

\bibitem{sanghi2023sketch}
Aditya Sanghi, Pradeep~Kumar Jayaraman, Arianna Rampini, Joseph Lambourne,
  Hooman Shayani, Evan Atherton, and Saeid~Asgari Taghanaki.
\newblock Sketch-a-shape: Zero-shot sketch-to-3d shape generation.
\newblock {\em arXiv preprint arXiv:2307.03869}, 2023.

\bibitem{sargent2023vq3d}
Kyle Sargent, Jing~Yu Koh, Han Zhang, Huiwen Chang, Charles Herrmann, Pratul
  Srinivasan, Jiajun Wu, and Deqing Sun.
\newblock Vq3d: Learning a 3d-aware generative model on imagenet.
\newblock {\em arXiv preprint arXiv:2302.06833}, 2023.

\bibitem{sauer2021projected}
Axel Sauer, Kashyap Chitta, Jens M{\"u}ller, and Andreas Geiger.
\newblock Projected gans converge faster.
\newblock {\em Advances in Neural Information Processing Systems (NeurIPS)},
  34:17480--17492, 2021.

\bibitem{takida2022sq}
Yuhta Takida, Takashi Shibuya, Weihsiang Liao, Chieh-Hsin Lai, Junki Ohmura,
  Toshimitsu Uesaka, Naoki Murata, Shusuke Takahashi, Toshiyuki Kumakura, and
  Yuki Mitsufuji.
\newblock Sq-vae: Variational bayes on discrete representation with
  self-annealed stochastic quantization.
\newblock In {\em International Conference on Machine Learning (ICML)}, pages
  20987--21012. PMLR, 2022.

\bibitem{van2017neural}
Aaron Van Den~Oord, Oriol Vinyals, et~al.
\newblock Neural discrete representation learning.
\newblock In {\em Proceedings of the 31st International Conference on Neural
  Information Processing Systems (NeurIPS)}, 2017.

\bibitem{vuong2023vector}
Tung-Long Vuong, Trung Le, He Zhao, Chuanxia Zheng, Mehrtash Harandi, Jianfei
  Cai, and Dinh Phung.
\newblock Vector quantized wasserstein auto-encoder.
\newblock {\em arXiv preprint arXiv:2302.05917}, 2023.

\bibitem{williams2020hierarchical}
Will Williams, Sam Ringer, Tom Ash, David MacLeod, Jamie Dougherty, and John
  Hughes.
\newblock Hierarchical quantized autoencoders.
\newblock In {\em Advances in Neural Information Processing Systems (NeurIPS)},
  volume~33, 2020.

\bibitem{wu2022nuwa}
Chenfei Wu, Jian Liang, Lei Ji, Fan Yang, Yuejian Fang, Daxin Jiang, and Nan
  Duan.
\newblock N{\"u}wa: Visual synthesis pre-training for neural visual world
  creation.
\newblock In {\em Computer Vision--ECCV 2022: 17th European Conference, Tel
  Aviv, Israel, October 23--27, 2022, Proceedings, Part XVI}, pages 720--736.
  Springer, 2022.

\bibitem{xiao2017online}
Han Xiao, Kashif Rasul, and Roland Vollgraf.
\newblock Fashion-mnist: a novel image dataset for benchmarking machine
  learning algorithms.
\newblock {\em arXiv}, 2017.

\bibitem{yan2021videogpt}
Wilson Yan, Yunzhi Zhang, Pieter Abbeel, and Aravind Srinivas.
\newblock Videogpt: Video generation using vq-vae and transformers.
\newblock {\em arXiv preprint arXiv:2104.10157}, 2021.

\bibitem{yu2015lsun}
Fisher Yu, Ari Seff, Yinda Zhang, Shuran Song, Thomas Funkhouser, and Jianxiong
  Xiao.
\newblock Lsun: Construction of a large-scale image dataset using deep learning
  with humans in the loop.
\newblock {\em arXiv preprint arXiv:1506.03365}, 2015.

\bibitem{yu2022vectorquantized}
Jiahui Yu, Xin Li, Jing~Yu Koh, Han Zhang, Ruoming Pang, James Qin, Alexander
  Ku, Yuanzhong Xu, Jason Baldridge, and Yonghui Wu.
\newblock Vector-quantized image modeling with improved {VQGAN}.
\newblock In {\em Proceedings of the International Conference on Learning
  Representations (ICLR)}, 2022.

\bibitem{zhang2018unreasonable}
Richard Zhang, Phillip Isola, Alexei~A Efros, Eli Shechtman, and Oliver Wang.
\newblock The unreasonable effectiveness of deep features as a perceptual
  metric.
\newblock In {\em Proceedings of the IEEE Conference on Computer Vision and
  Pattern Recognition (CVPR)}, pages 586--595, 2018.

\bibitem{zheng2022high}
Chuanxia Zheng, Guoxian Song, Tat-Jen Cham, Jianfei Cai, Dinh Phung, and Linjie
  Luo.
\newblock High-quality pluralistic image completion via code shared vqgan.
\newblock {\em arXiv preprint arXiv:2204.01931}, 2022.

\bibitem{zhengmovq}
Chuanxia Zheng, Long~Tung Vuong, Jianfei Cai, and Dinh Phung.
\newblock Movq: Modulating quantized vectors for high-fidelity image
  generation.
\newblock In {\em Proceedings of the 36st International Conference on Neural
  Information Processing Systems (NeurIPS)}, 2022.

\end{thebibliography}
